%% file: main.tex
\documentclass[10pt,twocolumn,letterpaper]{article}
\usepackage[pagenumbers]{style}
\input{preamble}

\definecolor{customblue}{rgb}{0.21,0.49,0.74}
\usepackage[pagebackref,breaklinks,colorlinks,allcolors=customblue]{hyperref}
\title{\textit{NAMD:} Virtual Follow-up Computed Tomography Synthesis via Nodule-Aligned Multimodal Diffusion Models for Early Lung Cancer Diagnosis}

\author{
James Song$^{1*}$ \quad Yifan Wang$^{1*}$ \quad Chuan Zhou$^{2}$ \quad Liyue Shen$^{1}$\\
$^1$University of Michigan \quad $^2$University of Michigan Medical School\\
{\tt\small \{shxjames,wangyfan,liyues\}@umich.edu, chuan@med.umich.edu}
}

\input{math_commands}

\begin{document}
\maketitle
\def\thefootnote{\fnsymbol{footnote}}
\footnotetext[1]{Equal contribution.}
\input{Sections/00_abstract}

\input{Sections/01_introduction}

\input{Sections/02_Related_Work}
\input{Sections/04_Method}

\input{Sections/05_Experiments}
\input{Sections/06_Conclusion}
\section*{Acknowledgment}
Chuan Zhou and Yifan Wang are supported in part by the National Institutes of Health grant number U01CA216459. 
Liyue Shen acknowledges funding support by NSF (National Science Foundation) via grant IIS-2435746, Defense Advanced Research Projects Agency (DARPA) under Contract No. HR00112520042, as well as the University of Michigan MICDE Catalyst Grant Award and MIDAS PODS Grant Award.
{
    \small
    \bibliographystyle{ieeenat_fullname}
    \bibliography{main}
}
\newpage
\appendix
\input{Sections/Appendix}

\end{document}

%% file: preamble.tex
\usepackage{pifont}

\usepackage{xcolor}
\usepackage{graphicx}
\usepackage{amsmath}
\usepackage{amssymb}
\usepackage{booktabs}
\usepackage{tcolorbox}
\usepackage{multirow}
\usepackage{bigdelim}

\definecolor{commentcol}{HTML}{C0392B}


%% file: math_commands.tex
\usepackage{amsmath,amsfonts,bm}

\def\eqref#1{equation~\ref{#1}}

\def\1{\bm{1}}

\def\ve{{\bm{e}}}
\def\vf{{\bm{f}}}

\def\vh{{\bm{h}}}

\def\vp{{\bm{p}}}

\def\vz{{\bm{z}}}

\def\mP{{\bm{P}}}
\def\mQ{{\bm{Q}}}

\def\mX{{\bm{X}}}

\DeclareMathAlphabet{\mathsfit}{\encodingdefault}{\sfdefault}{m}{sl}
\SetMathAlphabet{\mathsfit}{bold}{\encodingdefault}{\sfdefault}{bx}{n}

\newcommand{\KL}{D_{\mathrm{KL}}}



%% file: Sections/00_abstract.tex
\begin{abstract}
Lung cancer remains the leading cause of cancer-related mortality worldwide, with survival outcomes critically dependent on early and accurate detection. When low-dose computed tomography (LDCT) findings are indeterminate, clinicians typically defer diagnosis pending follow-up CT imaging obtained up to 12 months later, inevitably delaying treatment for patients with malignant nodules. To address this clinical gap, we propose \textbf{N}odule-\textbf{A}ligned \textbf{M}ultimodal (Latent) \textbf{D}iffusion (NAMD), a novel generative framework that synthesizes one-year follow-up nodule CT images conditioned on the baseline CT scan, quantitative nodule biomarkers, and patient-level Electronic Health Records (EHR), enabling timely prediction of nodule malignant progression without requiring actual follow-up scans. NAMD introduces two key contributions: (i) a \textbf{nodule-aligned latent space} regularized so that embedding distances reflect clinically meaningful biomarker changes, and (ii) an \textbf{LLM-driven multimodal conditioning mechanism} encoding heterogeneous EHR data into the diffusion backbone. Evaluated on the National Lung Screening Trial (NLST), our method's synthetic follow-up images achieve an AUROC of 0.805 and an AUPRC of 0.346 for lung nodule malignancy prediction, outperforming both the baseline LDCT performance without virtual follow-up generation, and existing state-of-the-art conditional generation methods, while maintaining competitive image quality. These findings suggest that NAMD enables earlier and more accurate lung cancer diagnosis by capturing clinically meaningful features of nodule progression.
\end{abstract}

%% file: Sections/01_introduction.tex
\section{Introduction}
\input{Figs/illustration/motivation}
Lung cancer, the leading cause of cancer-related mortality worldwide, has an overall 5-year relative survival rate of only 22.9 $\%$. Prognosis improves substantially with early detection: the 5-year survival rate reaches 61.2 $\%$ for patients diagnosed with localized tumors, compared to just 7$\%$ for advanced-stage disease~\citep{CancerStat, Cancer}. However, early detection of lung cancer remains particularly challenging, as an limited understanding of nodule progression and inherent biological uncertainty complicate the identification of early-stage malignant lesions~\citep{crosby2022early}. As a result, only 15$\%$ of lung cancer patients are diagnosed at an early stage~\citep{Assessment}. Low-dose computed tomography (LDCT) is widely used to screen high-risk populations for activate surveillance, but when its findings are indeterminate, radiologists typically recommend follow-up imaging up to twelve-month intervals to monitor nodule progression. 
During this period, patients with malignant nodules may face critical delays in definitive diagnosis and treatment.

With the rise of Artificial Intelligence (AI) for the healthcare, numerous machine learning~\citep{gupta2024texture,liu2024lung} and deep learning~\citep{yu2025etmo,ardila2019end,wang2024leveraging} models have been developed for lung cancer diagnosis, aiming to accurately classify lung nodules as malignant or benign. However, most existing approaches treat lung nodule diagnosis as a static, single time-point classification task. As shown in \cref{fig:motivation}, a nodule from baseline scan can appear benign while the one year follow-up reveals malignant progression. \textit{Generative models offer a way to anticipate this: by synthesizing follow-ups that approximate the real future scan, they can inform early diagnosis.}

Recent advances in conditional generative models have shown remarkable success in synthesizing high-fidelity images conditioned on semantic inputs such as text prompts~\citep{flux2024,Tan_2025_ICCV}. Prior studies~\citep{wang2024enhancing,liu2025imageflownet,wu2025early,chen2026learning} have explored generative models for disease progression prediction and reported promising results, offering a new avenue for achieving early diagnosis. However, clinical prognosis requires a level of precision beyond broad semantic conditioning~\citep{TangICCV2025TULIP,liu-etal-2025-medebench}. Specifically, accurate lung nodule progression prediction demands fine-grained control over both nodule- and patient-specific factors, ensuring that generated outcomes adhere to clinical biomarkers rather than loosely related semantics.
\vspace{-10pt}
\paragraph{Our Contributions.}
In this study, we propose \textbf{N}odule-\textbf{A}ligned \textbf{M}ultimodal (Latent) \textbf{D}iffusion (\textbf{\textit{NAMD}}) to model lung cancer progression by generating follow-up nodule images from baseline LDCT scans with quantitative nodule biomarkers and patients' Electronic Health Records (EHR), focusing on a fine-grained, controllable multimodal conditioning mechanism. 
The main contributions of this work are summarized as follows: 

\begin{itemize}
    \item We propose a \textbf{nodule-aligned latent space} 
    in which the geometry of latent embeddings is explicitly regularized to correspond to clinically meaningful changes in nodule biomarkers, thereby equipping the latent diffusion model with a structured representation of nodule progression across baseline and follow-up imaging.
    \item We propose an \textbf{LLM-driven diffusion conditioning mechanism} in which nodule- and patient-level metadata are first converted into structured radiology reports and subsequently encoded by a pretrained medical LLM with soft-prompt adaptation. 
    The resulting embeddings are injected into the diffusion backbone as auxiliary conditioning inputs, enabling fine-grained control over the generation of follow-up CT images.
    \item We evaluate NAMD on the NLST dataset~\citep{national2011national}, demonstrating strong performance in synthesizing one-year follow-up scans, and achieving improved diagnostic performance that outperforms baseline models, while nearly matching that of real follow-up scans. 
    These results validate NAMD's capability for early lung cancer diagnosis approximately one year in advance.
\end{itemize}

%% file: Figs/illustration/motivation.tex
\begin{figure}[t]
  \centering
  \includegraphics[width=0.8\columnwidth]{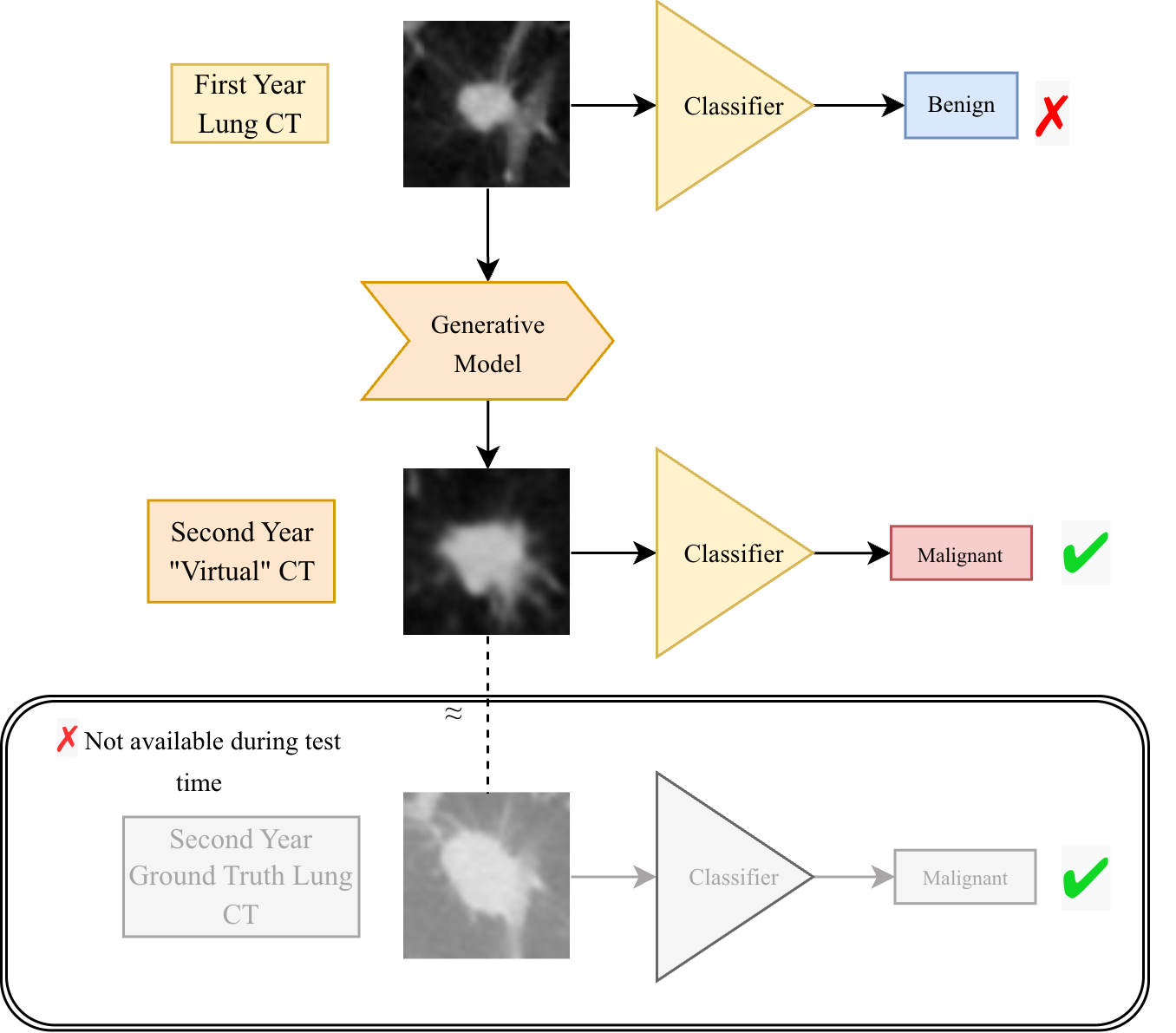}
  \caption{Illustration of how generative models can benefit early lung cancer diagnosis: generating a \emph{virtual} follow-up nodule CT image, which approximates the real future CT scan by predicting nodule progression, can improve the correct diagnosis from baseline CT image alone.}
  \vspace{-18pt}
  \label{fig:motivation}
\end{figure}

%% file: Sections/02_Related_Work.tex
\section{Related Works}
\subsection{Generative Models for Disease Progression} 
Generative models have emerged as a powerful tool for forecasting disease progression, with recent work favoring diffusion models for their high fidelity. 
Early works focused on improving generation image quality, such as the cascaded latent diffusion models (LDMs) for high-resolution chest X-ray synthesis~\cite{weber2023cascaded}. 
Building on this, several approaches have been proposed for disease trajectory prediction. For example, DDL-CXR \citep{yao2024addressing} utilizes LDMs to generate individualized chest X-rays for clinical prediction from asynchronous multi-modal data, and CXR-TFT~\citep{AroMeh_CXRTFT_MICCAI2025} introduces a multi-modal transformer to predict chest X-rays over time.
Recent works have explored alternatives of standard diffusion methods to better align with specific characteristics of diseases: ImageFlowNet~\citep{liu2025imageflownet} optimizes deterministic or stochastic flow fields within a representation space across patients and timepoints, and $\Delta-$LFM~\citep{chen2026learning} builds a patient-specific latent space by enforcing latents of MRIs from the same patient to lie on the same line in the latent space and using flow matching to model patient trajectories. 
Closest to our task on lung nodule progression prediction, McWGAN~\citep{wang2024enhancing} adopts a WGAN and CorrFlowNet~\citep{wu2025early} combines a correlational autoencoder with latent flow matching; however, both operate purely on imaging data, without integrating EHR information.

\subsection{Conditional Generation with Diffusion Models} 
Diffusion models \citep{sohldickstein2015deepunsupervisedlearningusing, ho2020denoisingdiffusionprobabilisticmodels} have become the dominant paradigm in conditional generations. Latent Diffusion Models (LDM)~\citep{Rombach_2022_CVPR} revolutionized high-resolution image synthesis by denoising in a compressed latent space, significantly reducing computational cost while maintaining perceptual quality. A key advantage of LDMs is their flexibility in incorporating conditional inputs of different modalities to guide generation. ControlNet~\citep{Zhang_2023_ICCV} and T2I-Adaptor~\citep{mou2023t2i} introduced efficient adaptation modules for more precise spatial control without altering pretrained weights. 
Recent advancements in Diffusion Transformers (DiTs)~\citep{Peebles_2023_ICCV} have shown excellent controllability, with methods like OmniControl~\citep{Tan_2025_ICCV} demonstrating universal control capabilities within transformer-based backbones.

Many existing conditional methods are designed to specify \textit{what} appears in an image via semantic and fine-grained spatial control. Modeling prognosis requires the subtle and continuous biological variations conditioned on individual clinical data. Our work targets this setting through a nodule-aligned latent space and an LLM-based conditioning mechanism, so generation is governed by patient-specific clinical attributes rather than image-level spatial guidance alone. 

%% file: Sections/04_Method.tex
\section{Method}
\input{Figs/LS}

We propose NAMD (\cref{fig:NAMD-Flowchart}), which adopts a LDM~\citep{Rombach_2022_CVPR} to generate ''virtual" one-year follow-up LDCT scans from baseline LDCT to support early lung cancer diagnosis. 
We design a nodule-aligned Variational Autoencoder (VAE) that compresses LDCT nodule images into compact latent embeddings. 
Within this latent space, a latent diffusion model, using LLM-constructed conditioning, models lung nodule progression. Unlike standard LDMs targeting broad semantic plausibility, NAMD requires generated progressions align with patient-specific EHR and nodule biomarkers.

\subsection{Notations}
\label{sec3.1:notations}
Let $(\mX^{(1)}, \mX^{(2)}, \ve, L) \sim \mathcal{D}$ denote a pair of longitudinal LDCT scans of the same lung nodule at two time points. Here, $\mX^{(1)}, \mX^{(2)} \in \mathbb{R}^{H \times W}$ are cropped, nodule-centered 2D images from the same patient's baseline scan and corresponding one-year follow-up scan, respectively. $\ve \in \mathbb{R}^d$ denotes the EHR vector containing nodule's biomarker attributes $\vf \in \mathbb{R}^T$ (e.g. diameter) corresponding to $\mX^{(1)}$, and the patient's EHR information $\vp \in \mathbb R^{d - T}$ (e.g. family cancer history).$L \in \{0, 1\}$ indicates the prediction target label of nodule malignancy. 
Our target problem is to model a conditional distribution $p(\mX^{(2)} | \mX^{(1)}, \ve)$ for generating follow-up lung nodule image. While fidelity to the ground truth $\mX^{(2)}$ is important, the ultimate goal is \textbf{downstream malignancy prediction} for early lung cancer diagnosis. 

\subsection{Nodule-Aligned Latent Space}
\label{sec:latent}
\input{Figs/alignment}
Recent work on Representational Autoencoders (RAE) \citep{zheng2026diffusion} shows that semantically rich spaces yield not only better reconstruction quality, but also superior generative quality in downstream diffusion models. However, standard VAEs \citep{DBLP:journals/corr/KingmaW13} map images into a lower-dimensional latent space without enforcing such structures, leaving the latent embeddings semantically unconstrained. One way to impose structure is supervised contrastive learning \citep{supcon}, which pulls and pushes samples based on discrete class labels. Our supervision, in contrast, is a continuous nodule feature vector $\vf$. Therefore, we introduce a \textit{nodule-aligned} latent space, in which a VAE is trained under two complementary objectives: a \textit{latent alignment loss}, which aligns latent neighbor structure with EHR-derived nodule biomarker similarity, and a \textit{predictive representation loss}, which guides the latent space using ground-truth malignancy labels.
\paragraph{Latent Alignment Loss} 
Suppose $i,j \in [B]$ be the index of two arbitrary data samples, where $B \in \mathbb{Z}^{+}$ denotes the batch size. 
Let $\vf_i \in \mathbb{R}^T$ be the feature vector containing the nodule EHR. Let $\vz_i = E(\mX_i)$ be the latent embedding of the LDCT scan $\mX_i$ encoded by the encoder $E$ of the VAE model. Denote the feature-space $(\vf)$ similarity and latent-space $(\vz)$ similarity between two data samples $(i,j)$ as
\begin{equation}
    \ell_{ij}^{(\vf)} =  -\frac{||\vf_j - \vf_i||^2}{2 \sigma_{\vf}^2}, \quad \ell_{ij}^{(\vz)} =  -\frac{||\vz_j - \vz_i||^2}{2 \sigma_{\vz}^2}
\end{equation}
where $\sigma_{\vf}, \sigma_{\vz} > 0$ are hyperparameters to control the sensitivity of the distance metric. Then, for each $i \neq j$, define the normalized rows as:
\begin{equation}
P_{ij} = \frac{\exp(\ell_{ij}^{(\vf)})}{\sum_{k \neq i} \exp(\ell_{ik}^{(\vf)})},
\qquad
Q_{ij} = \frac{\exp(\ell_{ij}^{(\vz)})}{\sum_{k \neq i} \exp(\ell_{ik}^{(\vz)})}
\label{eq:PiQidef}
\end{equation}
and $P_{ii} = Q_{ii} = 0$. In \cref{eq:PiQidef}, the numerator $\exp(\ell_{ij}^{(\vf)})$ and $ \exp(\ell_{ij}^{(\vz)})$ are large when samples $i, j$ are close. Since $\sum_{j \neq i} P_{ij} = \sum_{j \neq i} Q_{ij} = 1$, we have essentially constructed a neighbor distribution over relative similarities within the batch $B$. Finally, we introduce the Kullback-Liebler (KL) Divergence loss for the distribution matching:
\begin{equation}
\begin{aligned}
    \mP_i &= \text{Categorical}(P_{i1}, \dots, P_{iB})\\ \mQ_i &= \text{Categorical}(Q_{i1}, \dots, Q_{iB}) \\[1ex]
    \mathcal{L}_{\text{align}} &= \frac{1}{B} \sum_{i=1}^B \KL ( \mP_i \Vert \mQ_i ) \\ &= \frac{1}{B} \sum_{i=1}^B \sum_{j \neq i} P_{ij} \log \frac{P_{ij}}{Q_{ij}}.
\end{aligned}
\label{eq:align_loss}
\end{equation}
Minimizing the KL divergence in \cref{eq:align_loss} drives the latent-space neighborhood distribution $\mQ_i$ towards the feature-space neighborhood distribution $\mP_i$ as illustrated in \cref{fig:alignment}, so that nodules with similar attributes are embedded nearby, and dissimilar ones are pushed apart. 
Thus, the latent space geometry is shaped to mirror clinically meaningful similarity in nodule attributes. 

\paragraph{Predictive Representation Loss} 
Inspired by DDL-CXR~\citep{yao2024addressing}, we integrate \textbf{representation learning} with LDM training to ensure that the latent space captures clinical features relevant to the ultimate diagnostic task. We leverage the ground-truth malignancy label to structure the latent space by introducing a lightweight binary classification (benign v.s. malignant) objective $\mathcal{L}_{\text{pred}}$ on the latent $\vz$. We initialize a linear probe $h_\theta$ and optimize:
\begin{equation}
    \mathcal{L}_{\text{pred}} = -L \log(h_\theta(\vz)) - (1-L) \log(1-h_\theta(\vz)).
    \label{eq:pred_loss}
\end{equation}

\paragraph{Total Loss} 
Building upon standard VAE training strategy~\citep{DBLP:journals/corr/KingmaW13,Rombach_2022_CVPR}, we incorporate a weighted combination of L1 Reconstruction Loss $\mathcal{L_{\text{rec}}}$, a KL divergence loss to regularize learned latent embeddings towards a standard normal distribution $\mathcal{L}_{\text{KL}}$, a perceptual loss $\mathcal{L}_{\text{LPIPS}}$~\citep{zhang2018perceptual} to balance perceptual semantics and pixel-wise accuracy, our proposed alignment loss $\mathcal{L}_{\text{align}}$ (\cref{eq:align_loss}), and representation prediction loss $\mathcal{L}_{\text{pred}}$ (\cref{eq:pred_loss}):
\begin{equation}
\begin{aligned}
\mathcal{L}_{\text{VAE}}(\theta) ={}& \mathcal{L}_{\text{rec}}
+ \lambda_{KL} \mathcal{L}_{\text{KL}}
+ \lambda_{\text{LPIPS}} \mathcal{L}_{\text{LPIPS}} \\
&+ \lambda_{\text{align}} \mathcal{L}_{\text{align}}
+ \lambda_{\text{pred}} \mathcal{L}_{\text{pred}}
\end{aligned}
\label{eq:total_loss}
\end{equation}
where $\lambda_{KL}, \lambda_{\text{LPIPS}},\lambda_{\text{align}}, \lambda_{\text{pred}}$ denotes the hyperparameters of weighting for each loss.

\subsection{LLM-driven Multimodal Conditioning in LDM}
Because patients with morphologically similar baselines can follow divergent progression paths depending on clinical factors (e.g. age, family history), conditioning on EHR is necessary to produce follow-up scans personalized to the individual. Encoding the patient's EHR $\ve$ for conditioning the diffusion process requires a representation that captures clinically relevant and heterogeneous information for diffusion conditioning mechanism. 
While CLIP- and BERT-style encoders \citep{pmlr-v139-radford21a, devlin-etal-2019-bert, wang-etal-2022-medclip} are well-suited for broad semantic alignment, our setting requires a conditioning mechanism with fine-grained and precise controllability based on nodule biomarkers and other clinical values (e.g. nodule diameter, patient age).
LLMs have been shown to synthesise such fine-grained, compositional information more effectively when conditioning generative models~\cite{hu2024ella,yang2024mastering}. \citet{oh2024llm} further shows that an LLM adapted via soft-prompt tuning can produce such representations, cross-referencing clinical text with image features, without the cost of full model finetuning. Motivated by these observations, we propose an LLM-driven conditioning mechanism that renders the EHR $\ve$ as a structured radiology report and embeds it via a soft-prompt-adapted medical LLM, producing a fine-grained representation for diffusion conditioning.


\paragraph{LLM Adaptation with Learnable Prompts.} 
We employ MedGemma 1.5 (4B)~\citep{sellergren2025medgemma} as the backbone of medical LLM, which is specifically designed for medical report understanding. 
To efficiently adapt the model under limited downstream data, we adopt a soft-prompt post-training adaptation paradigm~\citep{oh2024llm}, where a set of learnable prompts are trained to enable task-specific adaptation while keeping the original model weights frozen.

Given the EHR $\ve \in \mathbb R ^ d$, we first convert it into a structured radiology report format (see \cref{fig:NAMD-Flowchart}) and then encode it into the LLM embedding space by using the adapted LLM model, yielding $\mathrm{emb}(\ve) \in \mathbb{R}^{N \times D}$ with sequence length $N$. We prepend $m$ sets of learnable soft prompts $\mathbf{s}^{(1)}, \mathbf{s}^{(2)}, \dots, \mathbf{s}^{(m)} \in \mathbb{R}^{n \times D}$ of sequence length $n$ to the embedding  $\mathrm{emb}(\ve)$, forming $m$ sets of input embedding sequences.  In other words, the text prompt embeddings are constructed as:
\begin{equation}
\begin{aligned}
        \mathbf{s}^{(j)} = [s_{1}^{(j)}, \dots, s_{n}^{(j)}] \in \mathbb{R}^{n \times D}, s_{i}^{(j)} \in \mathbb{R}^D\\
    \mathbf{c}^{(j)} = \text{concat}(\mathbf{s}^{(j)}, \text{emb}(\ve), \text{emb}(\texttt{<EOS>}))
\end{aligned}
\label{eq:txt_prompt_emb}
\end{equation}
We then extract the hidden state representation corresponding to the \texttt{<EOS>} token: $\vh_j^{\texttt{<EOS>}} = \text{LLM}(\mathbf{c}^{(j)})_{[-1]}$, which serves as our axuiliary context embedding. Under MedGemma's causal attention it is the only token attending to the full sequence. Finally, we concatenate all $\vh_j^{<EOS>}$ to form a sequence:
\begin{equation}
    \mathbf{C} = [\vh_1^{\texttt{<EOS>}}, \dots, \vh_m^{\texttt{<EOS>}}]
\end{equation}
We assume these concatenated embeddings carry the contextual information of each clinical attribute. Stacking them across the sequence yields a compact representation of the full patient context, which the diffusion model attends to via cross attention. 

\subsection{Latent Diffusion Model Training}
In the nodule-aligned latent space, temporal evolution of latent states is modeled using a latent diffusion generation process. We propose a multimodal latent diffusion model built on a U-Net backbone. We adopt a two-stage training strategy: an \emph{unconditional} stage that learns the general latent distribution of high-quality lung LDCT images, and a \emph{conditional} stage that learns nodule progression prediction for synthesizing the follow-up LDCT image conditioned on the baseline image, nodule biomarker, and patient EHR. 
\paragraph{Unconditional Diffusion Training}

We first pre-train the denoising U-Net on single-timepoint latents to capture the distribution of lung nodule images. 
Given a latent embedding $\vz = \mathcal{E}(\mX)$, the forward diffusion process adds Gaussian noise over $T$ timesteps: $\vz_t = \sqrt{\bar\alpha_t}\,\vz + \sqrt{1 - \bar\alpha_t}\,\boldsymbol{\epsilon}$, where $\boldsymbol{\epsilon} \sim \mathcal{N}(\mathbf{0}, \mathbf{I})$ and $\bar\alpha_t$ follows the standard noise schedule~\citep{ho2020denoisingdiffusionprobabilisticmodels}. The unconditional objective trains the network $\boldsymbol{\epsilon}_\theta$ to predict the added noise signals:
\begin{equation}
    \mathcal{L}_{\text{uncond}} = \mathbb{E}_{\vz,\,\boldsymbol{\epsilon} \sim \mathcal{N}(\mathbf{0},\mathbf{I}),\, t} \Big[\big\| \boldsymbol{\epsilon} - \boldsymbol{\epsilon}_\theta(\vz_t,\, t) \big\|^2 \Big]
    \label{eq:ldm_uncond}
\end{equation}

\paragraph{Conditional Diffusion Training.}
Building on the pre-trained diffusion model weights from the first stage, we fine-tune the U-Net to model the conditional follow-up image generation $p(\vz^{(2)} \mid \vz^{(1)}, \ve)$. Given a longitudinal pair $(\mX^{(1)}, \mX^{(2)})$, the baseline scan is encoded to $\vz^{(1)} = \mathcal{E}(\mX^{(1)})$ and the follow-up latent $\vz^{(2)} = \mathcal{E}(\mX^{(2)})$ serves as the diffusion target. The noisy follow-up latent $\vz_t^{(2)}$ is constructed via the forward process on $\vz^{(2)}$. The baseline latent $\vz^{(1)}$ is channel-concatenated with $\vz_t^{(2)}$ as input to the U-Net, while the LLM-derived context embedding $\mathbf{C}$ is injected via cross-attention at every UNet layer~\citep{Rombach_2022_CVPR}. 

\begin{equation}
    \mathcal{L}_{\text{cond}} = \mathbb{E}_{\vz^{(1)}, \vz^{(2)},\, \boldsymbol{\epsilon},\, t} \Big[\big\| \boldsymbol{\epsilon} - \boldsymbol{\epsilon}_\theta\!\big(\vz^{(1)},\, \vz_t^{(2)},\, t,\, \mathbf{C}\big) \big\|^2 \Big]
    \label{eq:ldm_cond}
\end{equation}

At inference, given a new baseline scan $\mX^{(1)}$ and its associated EHR context, we sample $\vz_T^{(2)} \sim \mathcal{N}(\mathbf{0}, \mathbf{I})$ and iteratively denoise using $\boldsymbol{\epsilon}_\theta$ conditioned on $\vz^{(1)} = \mathcal{E}(\mX^{(1)})$ and $\mathbf{C}$ via DDIM~\citep{song2021denoising} with 50 steps, then decode the result via the VAE decoder $D$ to produce the predicted follow-up image $\hat{\mX}^{(2)} = D(\hat{\vz}^{(2)})$.

%% file: Figs/LS.tex
\begin{figure*}[ht]
    \centering
    \includegraphics[width=0.85\linewidth]{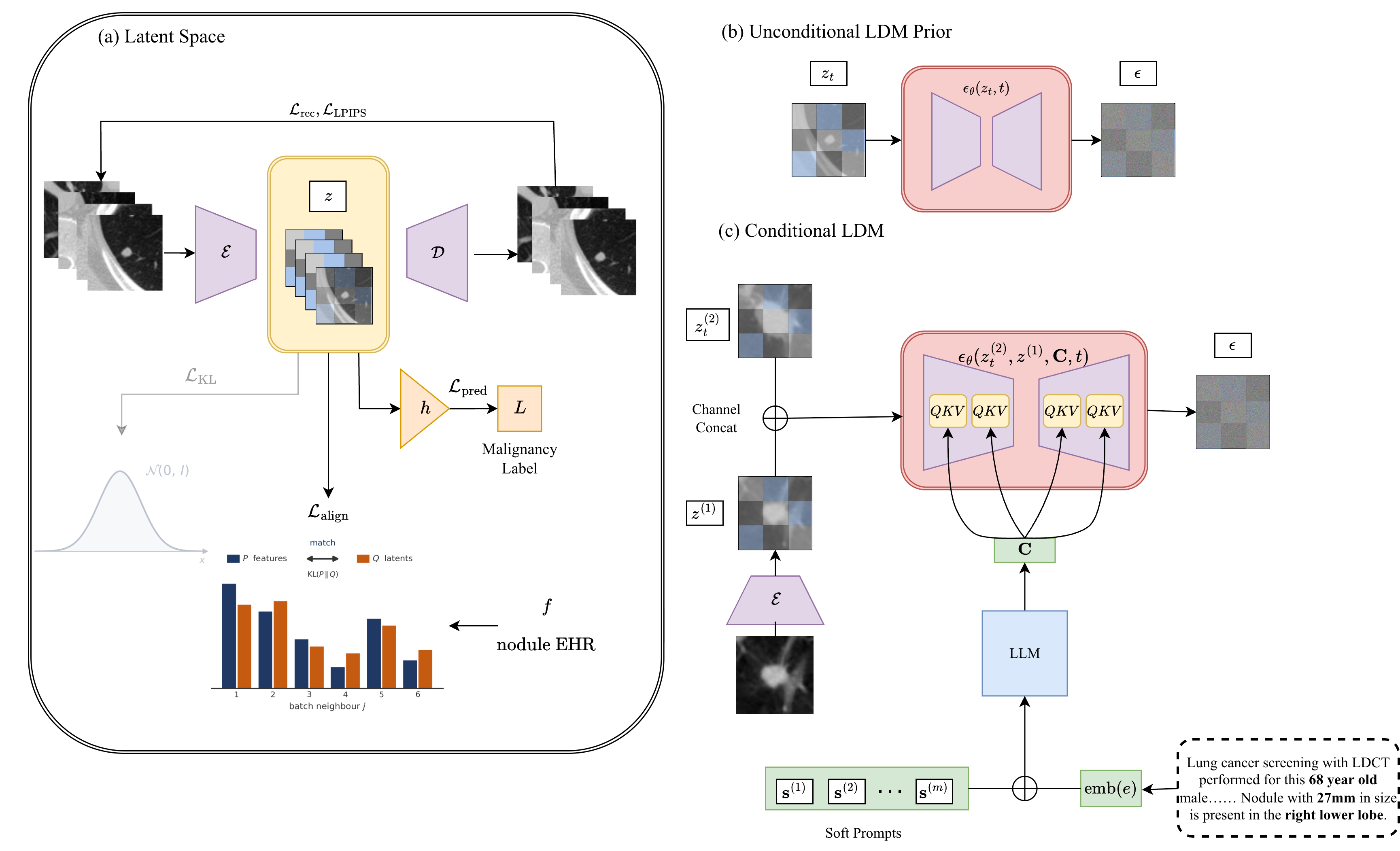}
    \caption{\textbf{NAMD framework.}
\textbf{(a)} A VAE ($\mathcal{E}$/$\mathcal{D}$) is trained with reconstruction and perceptual
losses. Latents $\vz$ are shaped via $\mathcal{L}_{\mathrm{align}}$ with a distribution matching objective and a malignancy-predictive head $\mathcal{L}_\mathrm{pred}$, in addition to a light KL regularization loss.
\textbf{(b)} A diffusion UNet $\epsilon_\theta$ learns an unconditional latent prior for high-quality image generation. \textbf{(c)} The same backbone is conditioned on the baseline image latent via concatenation and an LLM-derived context $\mathbf{C}$ including patient EHR and soft prompts adaptation via cross-attention.}
\vspace{-8pt}
    \label{fig:NAMD-Flowchart}
\end{figure*}

%% file: Figs/alignment.tex
\begin{figure*}[t]
    \centering
    \includegraphics[width=0.8\linewidth]{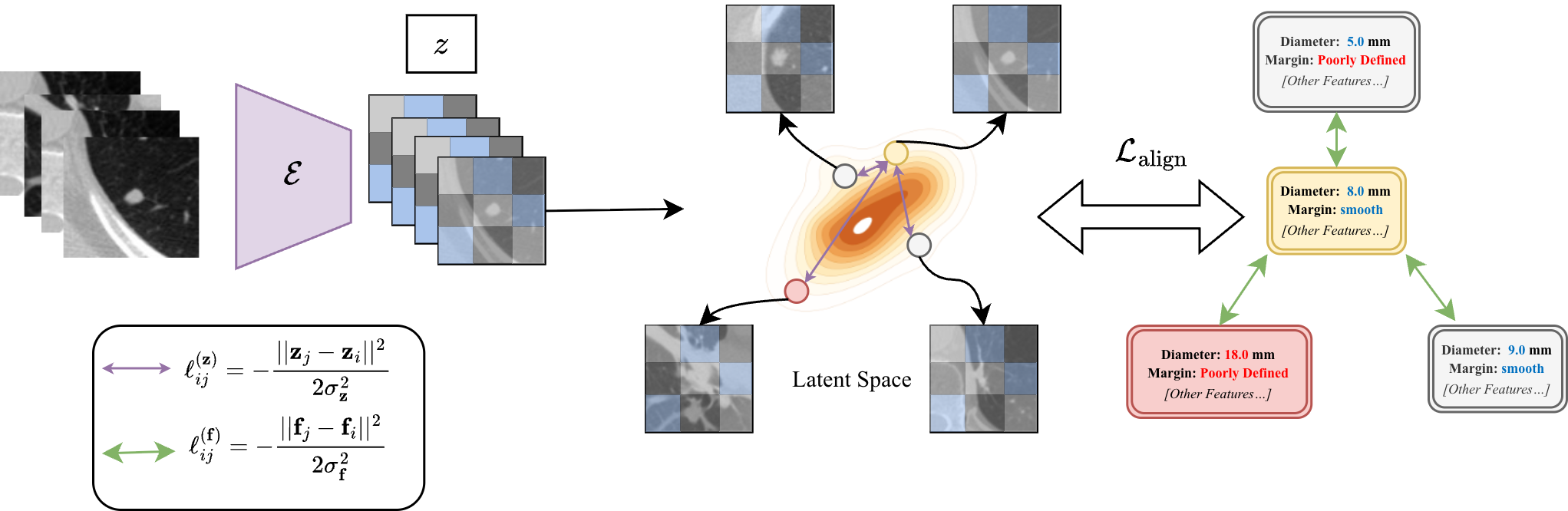}
    \caption{\textbf{Latent Alignment loss} in the nodule-aligned latent space: $\mathcal{L}_{\text{align}}$ matches pairwise similarities of latents constructed from $\ell_{ij}^{(\vz)}$ to pairwise similarities of nodule biomarker features constructed from $\ell_{ij}^{(\vf)}$ to organize the latent space based on nodule attributes.} 
    \vspace{-15pt}
    \label{fig:alignment}
\end{figure*}

%% file: Sections/05_Experiments.tex
\section{Experiments and Discussion}
\subsection{Experimental Setup}

\subsubsection{Dataset}
\label{dataset}
\paragraph{Longitudinal Nodule Progression Dataset.}
We study lung nodule progression for early diagnosis on the National Lung Screening Trial (NLST) dataset~\citep{national2011national}, selecting 1,226 subjects with clinically indeterminate nodules and at least one longitudinal follow-up. For our conditional LDM training (stage 3 in Fig~\ref{fig:NAMD-Flowchart}), we use 1,121 baseline--follow-up image pairs from 776 subjects (165 malignant, 611 benign). The remaining 450 subjects form an independent hold-out test set of 450 pairs (53 malignant, 397 benign). All splits are done at the patient level to prevent data leakage. 

\paragraph{Pretraining and Diagnosis Classifier Training Dataset.}
The autoencoder and the unconditional U-Net (stages 1--2 in Fig~\ref{fig:NAMD-Flowchart}), together with the diagnosis classifier with a vision transformer (ViT) backbone, are trained on single time-point LDCT images. 
We use a large cohort combining multiple datasets, including DLCS~\citep{wang2025duke}, LUNA16~\citep{setio2017validation}, LUNA25~\citep{peeters_2025_15094631}, and a subset of NLST, to construct a dataset with a total of 10610 images. 
This cohort shares no patients with the progression test set to avoid data leakage. The ViT classifier is trained only on real LDCT scans and never sees generated images, keeping it independent from the generation pipeline it evaluates.

\subsubsection{Evaluation Metrics}
We evaluate NAMD on three axes: diagnostic utility, nodule-attribute fidelity, and image quality. Diagnostic utility is our primary goal, but the virtual follow-up should still have the central clinical properties and be visually realistic. All metrics are reported as mean and standard deviation across $K=20$ independent runs; details on the choice of $K$ are provided in \cref{app:pred_var}.
\paragraph{Diagnostic Utility.} 
We pass generated follow-ups through a frozen ViT based binary classifier and report diagnosis performance including AUROC and AUPRC. The ViT is pre-trained only on the real single time-point LDCT cohort (Sec.~\ref{dataset}), which spans several datasets and shares no patients with the progression test set, so it cannot exploit dataset-specific shortcuts or NAMD's generation artifacts. All diagnostic results come from the held-out test set, unseen by both NAMD and the ViT.
\paragraph{Nodule-attribute Fidelity.} 
We also assess how well generated follow-ups reproduce the real one-year change in the nodule's clinical attributes. We segment the nodule mask from each generated follow-up with MedSAM~\citep{ma2024segment} and measure four clinical focused properties. Nodule size and Hounsfield Unit (HU) density change is measured by the Pearson correlation $r_{\mathrm{area}}, r_{\mathrm{HU}}$ between the generated and real nodule areas. 
Boundary regularity is measured by the mean absolute error in PyRadiomics 2-D Sphericity (Sphere MAE), a scale-invariant descriptor of margin smoothness that separates benign-typical from spiculated or lobulated malignancy-typical margins. Texture evolution is measured by the cosine similarity between $\Delta_{\mathrm{real}} = z(T_{1,\mathrm{real}}) - z(T_0)$ and $\Delta_{\mathrm{gen}} = z(T_{1,\mathrm{gen}}) - z(T_0)$ in the 93-dimensional $z$-scored IBSI radiomic feature space (Traj-cos), testing whether the generator captures the diagnostic baseline-to-follow-up trajectory. 
\paragraph{Image Quality.} We report FID and LPIPS~\citep{zhang2018perceptual} between generated and real follow-up images. Since the single observed follow-up is only one plausible future of a biologically uncertain process, we measure distributional realism and perceptual consistency rather than per-pixel accuracy.

\subsubsection{Implementation Details}
For image processing, each image is initially cropped to a size of $64\times64$ pixels, with the nodule in the center, followed by random rotations of $45^\circ, 90^\circ, 135^\circ, 180^\circ$  as a data augmentation technique to increase the amount of training samples. For VAE training (Section~\ref{sec:latent}), we follow equation~\ref{eq:total_loss}, and use $\lambda_{KL} = 10^{-6}, \lambda_{\text{LPIPS}} = 1, \lambda_{\text{align}} = 1, \lambda_{\text{pred}} = 0.2$. For the conditional LDM training, we set $n=8$ as the soft prompt context length and $m=2$ as the number of soft prompts (equation~\ref{eq:txt_prompt_emb}). For both VAE and UNet, we adopt pretrained SD1.5 weights~\citep{Rombach_2022_CVPR} due to the small scale of our dataset. We include ablation results against pretrained weights in \cref{app:pretrained_ablation}.

\subsubsection{Baseline Comparison}
We compare our method with several representative deterministic and stochastic generation approaches. We finetune from pretrained weights for SD1.5 \citep{Rombach_2022_CVPR}, which is a representative method in text-to-image methods. We evaluate against McWGAN \citep{wang2024enhancing} and CorrFlowNet~\citep{wu2025early} which target the same progression prediction task we do. As DDL-CXR~\citep{yao2024addressing} and ImageFlowNet (SDE)~\citep{liu2025imageflownet} were originally developed for domains distinct from lung LDCT scans and utilize specialized architectures, we trained these models from scratch following their original implementations.

\subsection{Main Results}

\input{Tables/benchmark}

\paragraph{Diagnostic Performance.} 
As demonstrated in \cref{tab:combined_metrics}, NAMD substantially improves diagnostic performance by predicting follow-up nodules from baseline data. The resulting diagnosis model achieved a test AUROC of $0.805$ and an AUPRC of $0.346$ when using NAMD-generated follow-up nodule images in settings where only real baseline information is available. This represents a substantial gain over real baseline images (AUROC: $0.742$) and approaches the performance based on actual follow-up nodules (AUROC: $0.819$), effectively diminishing the gap between baseline and future clinical data. NAMD effectively synthesizes follow-up images that preserve clinically relevant features for malignancy assessment. When compared with baseline models, our method outperforms all baseline approaches in both AUROC and AUPRC. 

\paragraph{Nodule Attributes.} 
NAMD achieves the best $r_{\text{HU}}$ (0.722) and ties for second best Sphere MAE (0.102) across all methods, while staying within 0.02 for $r_{\text{area}}$ and second best on Traj-cos, indicating that its generated nodules have largely preserved relevant nodule characteristics. McWGAN leads on Traj-cos and CorrFlowNet-ODE on $r_\text{area}$, both of which are deterministic methods. We attribute this to the nature of deterministic methods: rather than modeling a distribution over plausible trajectories a nodule may follow, they regress to a single outcome that discards patient-level variability. ImageFlowNet (IFN) leads on Sphere MAE by a tiny margin of $0.006$ mm, as it operates directly in pixel space and optimizes for mean pixel accuracy. However, this does not translate to realistic generation: its FID is worse by a wide margin (196.944 vs. 82.973 for NAMD), indicating substantially less realistic outputs.
\paragraph{Image Metrics}
NAMD achieves the second-best FID overall with LPIPS comparable to the strongest baselines. Interestingly, image quality and diagnostic utility are decoupled across the field for some methods: McWGAN attains the best FID ($80.44$) but only $0.307$ AUPRC, while ImageFlowNet records the worst FID ($196.944$) yet a competitive $0.335$ AUPRC. NAMD captures diagnostically relevant structure without sacrificing perceptual fidelity.

\subsection{Ablation Study}
\input{Figs/tsne}
\input{Tables/ablation_llm}
\input{Figs/variance}

\subsubsection{Latent Space.}
We provide visualizations of the learned latent space in \cref{fig:tsne} over the test set using t-SNE~\citep{JMLR:v9:vandermaaten08a}. NAMD constructs a latent space with structure as seen in \cref{fig:tsne} left: nodules with shorter diameters (dark purple) at the top transitions to nodules with increasingly longer diameters (green/yellow). This spatial distribution suggests that the NAMD autoencoder has successfully captured nodule diameter in its representation. In contrast, a VAE that is not nodule-aligned (\cref{fig:tsne} right) results in a more entangled representation space. We also fit a linear probing head on the latent space to recover longest diameter from latent $\vz$ with ridge regression. With alignment, the linear probe recovers longest diameter to $R^2 = 0.241$ with mean absolute error (MAE) $2.08$ mm compared to just $R^2 = -0.062$ and MAE $3.05$ mm for the unaligned counterpart, showing that the latent space has encoded diameter as a feature.

We also conducted an ablation study to further assess the contribution of the nodule-aligned latent space. As seen in \cref{tab:combined_metrics}, when those losses are omitted, AUROC and AUPRC drop from 0.805 to 0.779 and from 0.346 to 0.319 respectively. FID also worsens from 83.06 to 89.487, indicating a decrease in distribution similarity. All nodule-attribute metrics improved when the nodule-aligned latent space was used. While the unaligned variant achieves slightly improved LPIPS scores, this does not translate to either nodule attributes or improved diagnostic ability, indicating that structuring the latent space helps with capturing clinically meaningful nodule attributes. 

\subsubsection{LLM-driven EHR Conditioning}
\cref{tab:ehr-llm-ablation} isolates the contribution of each component in the LLM-EHR conditioning pathway, with all variants initialized from the same VAE and UNet prior from part (a) and (b). Encoding the EHR through an MLP yields a slight diagnostic gain over the image-only baseline from 0.758 to 0.760 AUROC, but replacing the MLP with a frozen MedGemma leads to an obvious improvement to 0.775. This shows that MedGemma, as an LLM, provides better encoding of information compared to a simpler trained EHR encoder. Moreover, adapting MedGemma via soft prompts (i.e. NAMD) brings AUROC to 0.805. This shows that the gain in diagnosis is due to how the EHR is encoded rather than just its inclusion. Notably, the diagnostic gains come at no perceptual cost: LPIPS remain effectively unchanged, whereas FID improves when LLM is included, suggesting that the LLM pathway injects clinically relevant conditioning signal without degrading image quality. 

\subsection{Prediction Variance }

To better understand how the generated images influence diagnostic confidence, we analyzed the relationship between prediction error ($|L - \bar{p}|$), predictive variance ($\text{Var}(p)$) (\cref{fig:prediction_variance}), and the spatial characteristics of the generated images (\cref{fig:qualitative_examples}). As shown in \cref{fig:error_vs_variance}, the model's output roughly fall into three categories: predictions with low error and variance (confidently correct), median error and high variance (unsure), and high error and low variance (confidently incorrect). In \cref{fig:variance_by_label}, we also observe that both benign and malignant cases have predictions with low and high variance, but malignant cases tend to have more confident predictions, whereas benign cases are more spread out in confidence. 

We also present qualitative assessments of the generated images in \cref{fig:qualitative_examples}. Notably, the variance concentrates predominantly along the boundary of the nodule, specifically the differences in nodule sizes and background. There are no clear distinction between the three categories in the mean error maps, indicating that pixel-level fidelity to the GT images does not fully determine downstream diagnostic accuracy. We conjecture that pixel-wise reconstruction and general perceptual performance does not capture all the factors for downstream clinical prediction. Instead, NAMD's latent space encodes clinically relevant representations, as see in  \cref{tab:combined_metrics} (nodule metrics) and the latent-geometry visualization (\cref{fig:tsne}). By encoding these clinically relevant progression signals rather than pixel-level detail, the generation task serves as an effective surrogate that captures disease trajectory, thereby improving early diagnostic.

%% file: Tables/benchmark.tex
\begin{table*}[t]
\centering
\fontsize{5.5}{8}\selectfont
\caption{Comparison of methods and results from ablation study in terms of image quality metrics, nodule-property fidelity, and diagnosis performance (AUROC and AUPRC). For stochastic methods, mean and standard deviation are calculated across 20 different samples from NAMD. \textbf{Bold} indicates best; \underline{underline} indicates second best.}
\label{tab:combined_metrics}
\begin{tabular}{l c c c c c c c c c}
\toprule
 & \multicolumn{2}{c}{\textbf{Image Quality}} & \multicolumn{4}{c}{\textbf{Nodule-attribute Fidelity}} & \multicolumn{2}{c}{\textbf{Diagnosis Performance}} & \\
 \cmidrule(lr){2-3} \cmidrule(lr){4-7} \cmidrule(lr){8-9} \cmidrule(lr){10-10}
\textbf{Method} & LPIPS $\downarrow$ & FID $\downarrow$ & $r_{\mathrm{area}}$ $\uparrow$ & $r_{\mathrm{HU}}$ $\uparrow$ & Sphere MAE $\downarrow$ & Traj-cos $\uparrow$ & AUROC $\uparrow$ & AUPRC $\uparrow$ & \textbf{Avg.\ Rank} $\downarrow$ \\
\midrule
\textit{Real Image Baselines} & & & & & & & & \\
Real baseline LDCT & - & - & - & - & - & - & 0.742 & 0.263 & - \\
Real follow-up LDCT & - & - & - & - & - & - & 0.819 & 0.393 & - \\
\hline
\textit{Deterministic Methods} & & & & & & & & \\
McWGAN \citep{wang2024enhancing} & 0.364 & \textbf{80.443} & 0.688 & \underline{0.716} & 0.104 & \textbf{0.439} & 0.763 & 0.307 & 4.25 \\
CorrFlowNet (ODE) \citep{wu2025early} & \textbf{0.202} & 91.695 & \textbf{0.723} & 0.566 & \underline{0.102} & 0.338 & \underline{0.779} & 0.318 & 4.00 \\
\hline
\textit{Stochastic Methods} & & & & & & & & \\
SD1.5 \citep{Rombach_2022_CVPR} & {0.207 $\pm$ 0.003} & 87.087 $\pm$ 0.800 & 0.551 $\pm$ 0.042 & 0.564 $\pm$ 0.028 & 0.132 $\pm$ 0.006 & 0.351 $\pm$ 0.015 & 0.701 $\pm$ 0.031 & 0.245 $\pm$ 0.032 & 6.13 \\
DDL-CXR \citep{yao2024addressing} & 0.235 $\pm$ 0.005 & 97.679 $\pm$ 1.954 & 0.551 $\pm$ 0.031 & 0.500 $\pm$ 0.040 & 0.136 $\pm$ 0.005 & 0.363 $\pm$ 0.014 & 0.704 $\pm$ 0.026 & 0.245 $\pm$ 0.029 & 7.00 \\
ImageFlowNet (SDE) \citep{liu2025imageflownet} & 0.337 $\pm$ 0.000 & 196.944 $\pm$ 1.131 & \underline{0.717 $\pm$ 0.009} & 0.538 $\pm$ 0.005 & \textbf{0.096 $\pm$ 0.002} & \underline{0.416 $\pm$ 0.005} & 0.771 $\pm$ 0.013 & \underline{0.335 $\pm$ 0.025} & 4.25 \\
CorrFlowNet (SDE) \citep{wu2025early} & \textbf{0.202 $\pm$ 0.001} & 88.560 $\pm$ 0.231 & 0.716 $\pm$ 0.004 & 0.569 $\pm$ 0.001 & \underline{0.102 $\pm$ 0.001} & 0.341 $\pm$ 0.004 & 0.776 $\pm$ 0.004 & 0.321 $\pm$ 0.011 & \underline{3.69} \\
\midrule
\textbf{NAMD (Ours)} & 0.220 $\pm$ 0.001 & \underline{82.973 $\pm$ 0.691} & 0.706 $\pm$ 0.017& \textbf{0.722 $\pm$ 0.010} & \underline{0.102 $\pm$ 0.003} & 0.412 $\pm$ 0.010 & \textbf{0.805 $\pm$ 0.018} & \textbf{0.347 $\pm$ 0.029} & \textbf{2.50} \\
NAMD (w/o nodule-aligned) & \underline{0.203 $\pm$ 0.002} & 89.487 $\pm$ 1.010 & 0.651 $\pm$ 0.028 & 0.696 $\pm$ 0.016 & 0.111 $\pm$ 0.003 & 0.405 $\pm$ 0.014 & \underline{0.779 $\pm$ 0.023} & 0.319 $\pm$ 0.039 & 4.19 \\
\bottomrule
\end{tabular}
\vspace{-12pt}
\end{table*}

%% file: Figs/tsne.tex

\begin{figure}[tbp]
    \centering
    \includegraphics[width=0.4\linewidth]{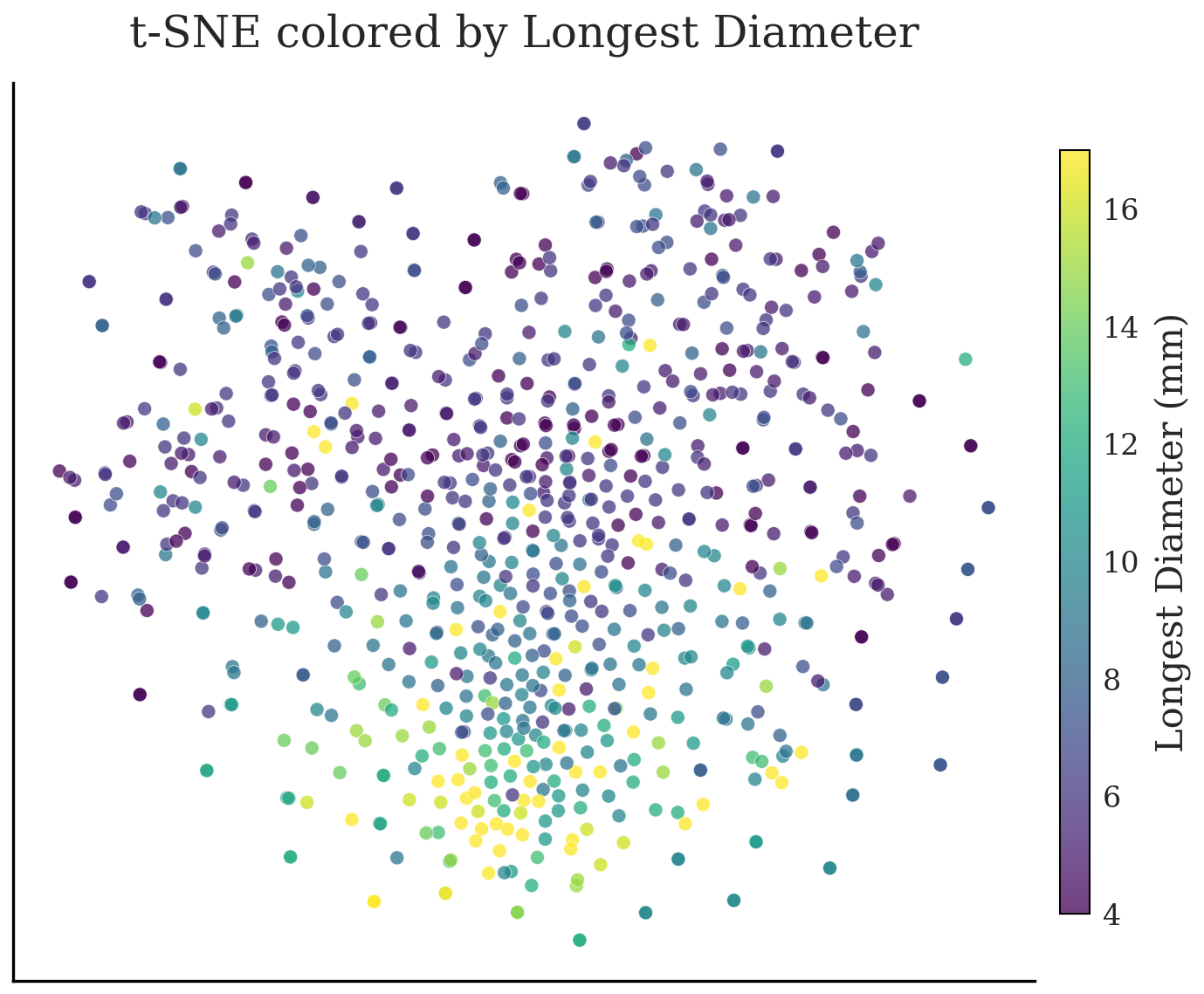}\hspace{2mm}
    \includegraphics[width=0.4\linewidth]{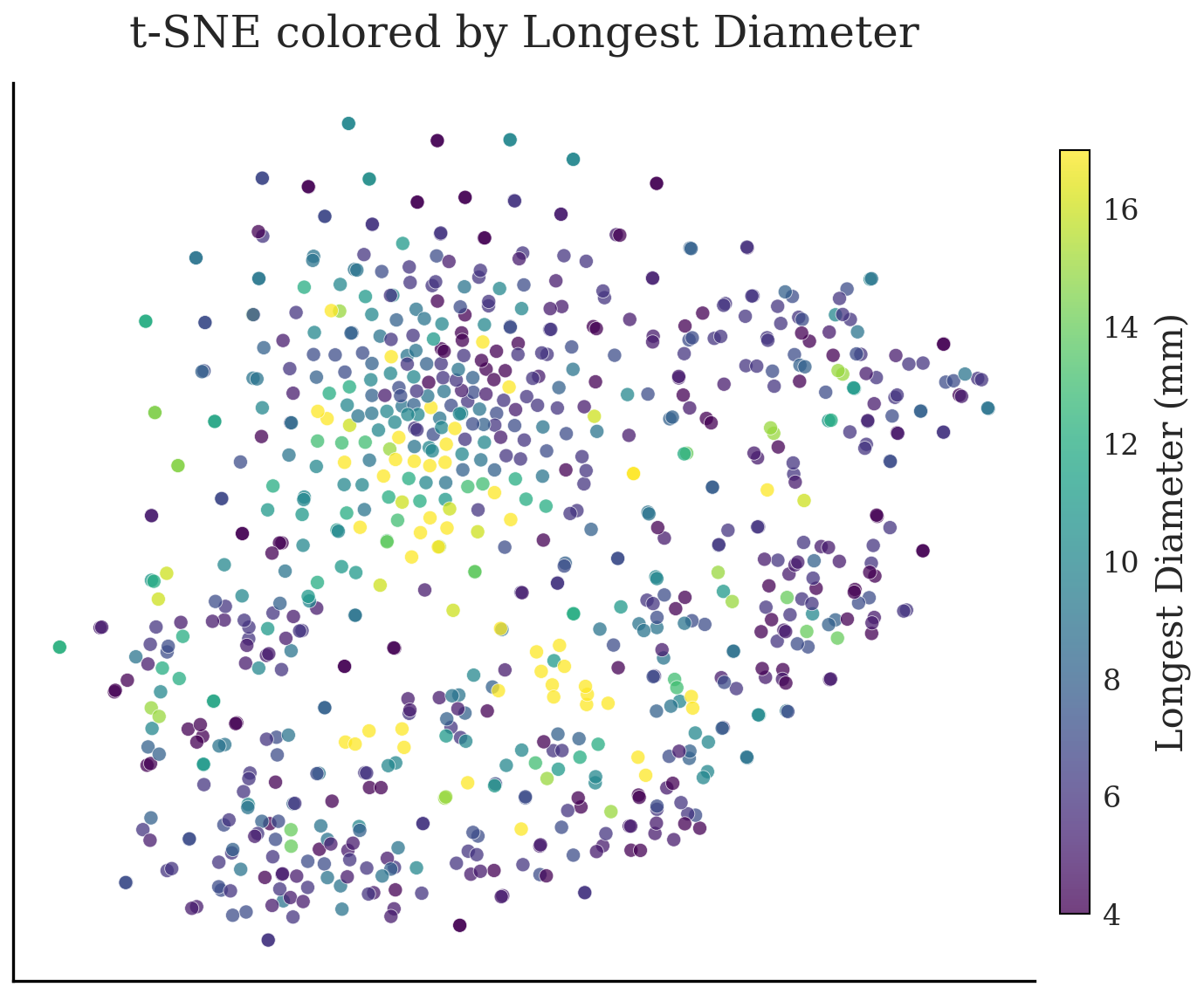}
    \caption{t-SNE of the learned latent space, \textbf{with} (left) and \textbf{without} (right) the alignment loss.}
    \label{fig:tsne}
    \vspace{-9pt}
\end{figure}

%% file: Tables/ablation_llm.tex
\begin{table}[t]
\centering
\fontsize{6}{8}\selectfont
\caption{Ablation over the EHR-LLM conditioning pathway. ``MLP'' encodes EHR with an MLP; ``LLM'' uses MedGemma; ``+SP'' adds soft-prompt tuning.}
\label{tab:ehr-llm-ablation}
\setlength{\tabcolsep}{4pt}
\begin{tabular}{lcccc}
\toprule
 & \multicolumn{2}{c}{Image Quality} & \multicolumn{2}{c}{Diagnosis} \\
\cmidrule(lr){2-3}\cmidrule(lr){4-5}
Conditioning & LPIPS $\downarrow$ & FID $\downarrow$ & AUROC $\uparrow$ & AUPRC $\uparrow$ \\
\midrule
Baseline (img only) & 0.222 $\pm$ 0.002 & 84.3 $\pm$ 1.14 & 0.758 $\pm$ 0.018 & 0.312 $\pm$ 0.029 \\
+ EHR (MLP)         & 0.222 $\pm$ 0.002 & 86.6 $\pm$ 0.91 & 0.760 $\pm$ 0.019 & 0.322 $\pm$ 0.045 \\
+ EHR (LLM)         & \textbf{0.218 $\pm$ 0.002} & 83.4 $\pm$ 0.99 & 0.775 $\pm$ 0.023 & 0.326 $\pm$ 0.039 \\
+ EHR (LLM) + SP    & 0.220 $\pm$ 0.001 & \textbf{83.0 $\pm$ 0.69} & \textbf{0.805 $\pm$ 0.018} & \textbf{0.347 $\pm$ 0.029} \\
\bottomrule
\end{tabular}
\end{table}

%% file: Figs/variance.tex
\begin{figure}[tb]
    \centering
    \begin{subfigure}{0.49\columnwidth}
        \centering
        \includegraphics[width=\linewidth]{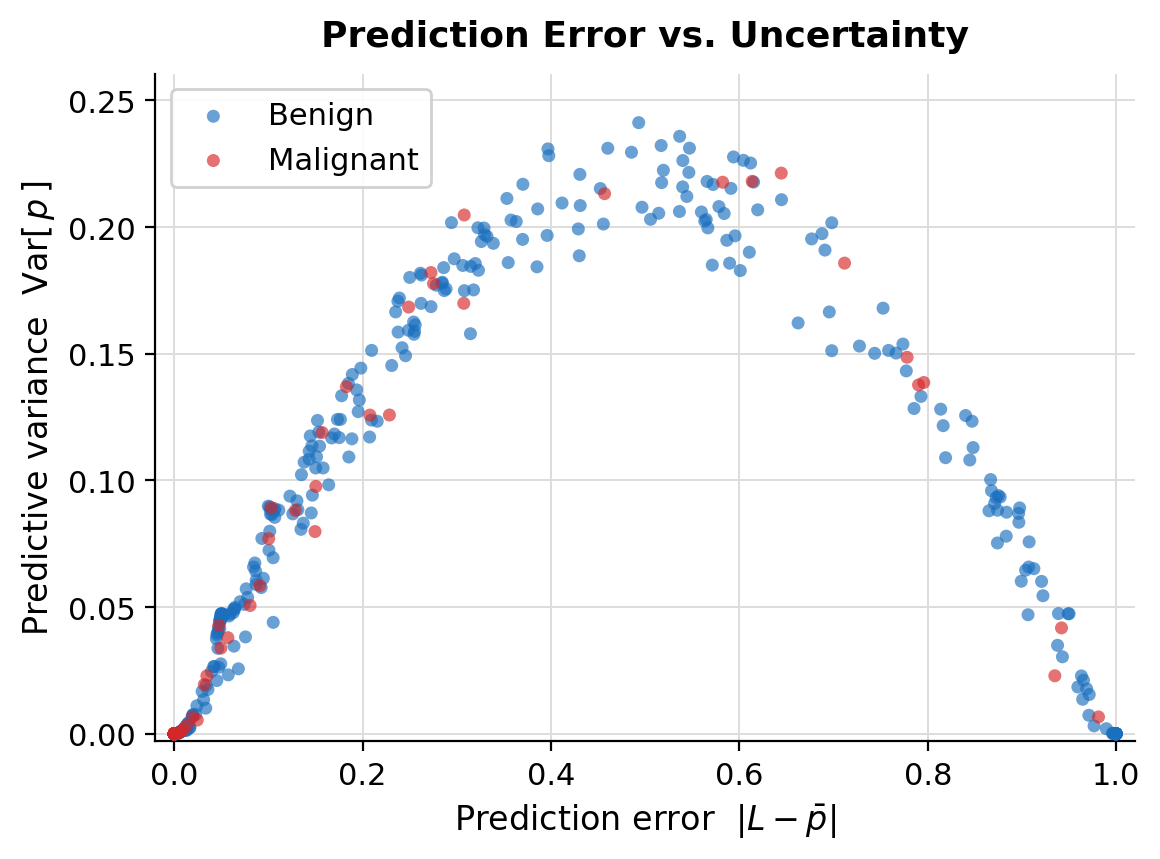}
        \caption{Prediction error $|L-\bar{p}|$ vs. predictive variance
    $\mathrm{Var}(p)$; $\bar{p}$ is the average probability across 20 runs.}
        \label{fig:error_vs_variance}
    \end{subfigure}
    \hfill
    \begin{subfigure}{0.49\columnwidth}
        \centering
        \includegraphics[width=\linewidth]{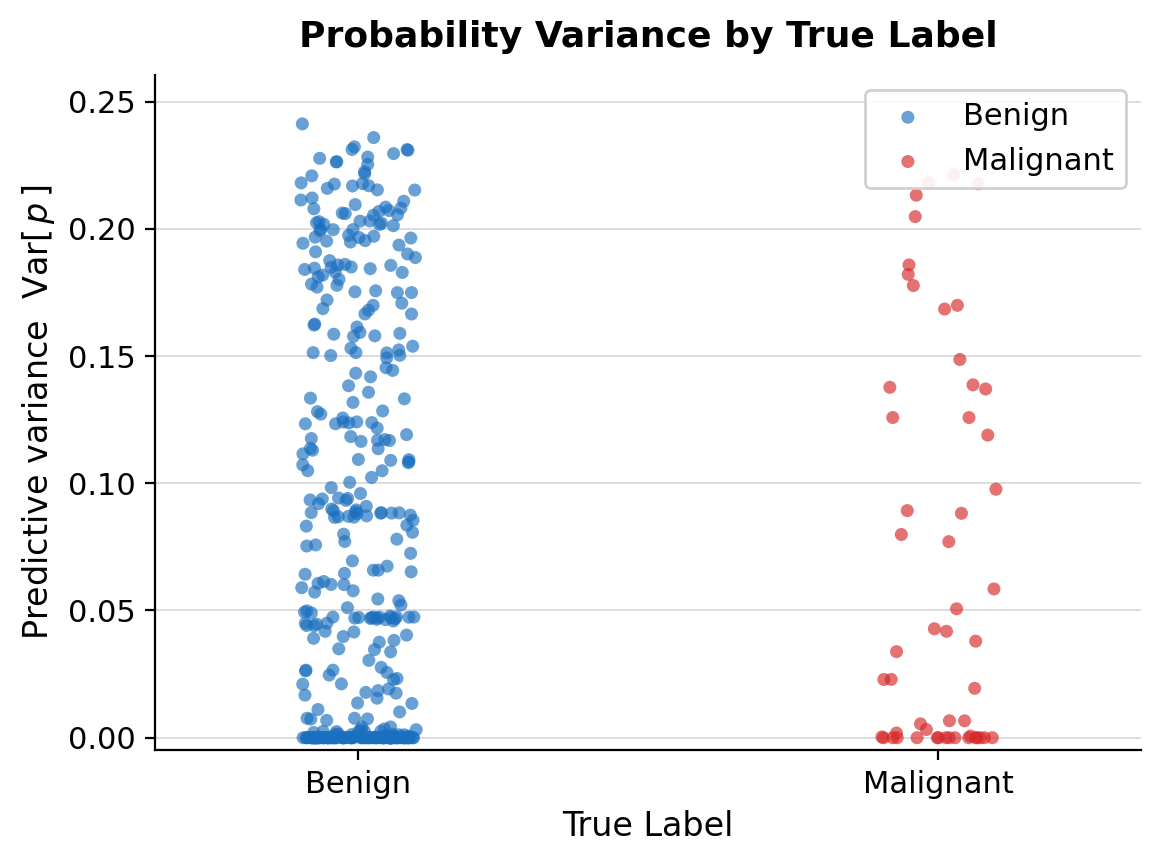}
        \caption{Variance by true label (Benign vs Malignant)}
        \label{fig:variance_by_label}
    \end{subfigure}
    \vspace{-5pt}
    \caption{Predictive uncertainty over the test set.}
    \label{fig:prediction_variance}
    \vspace{-10pt}
\end{figure}

\begin{figure*}[tb]
    \centering
    \vspace{-12pt}
    \includegraphics[width=0.8\textwidth]{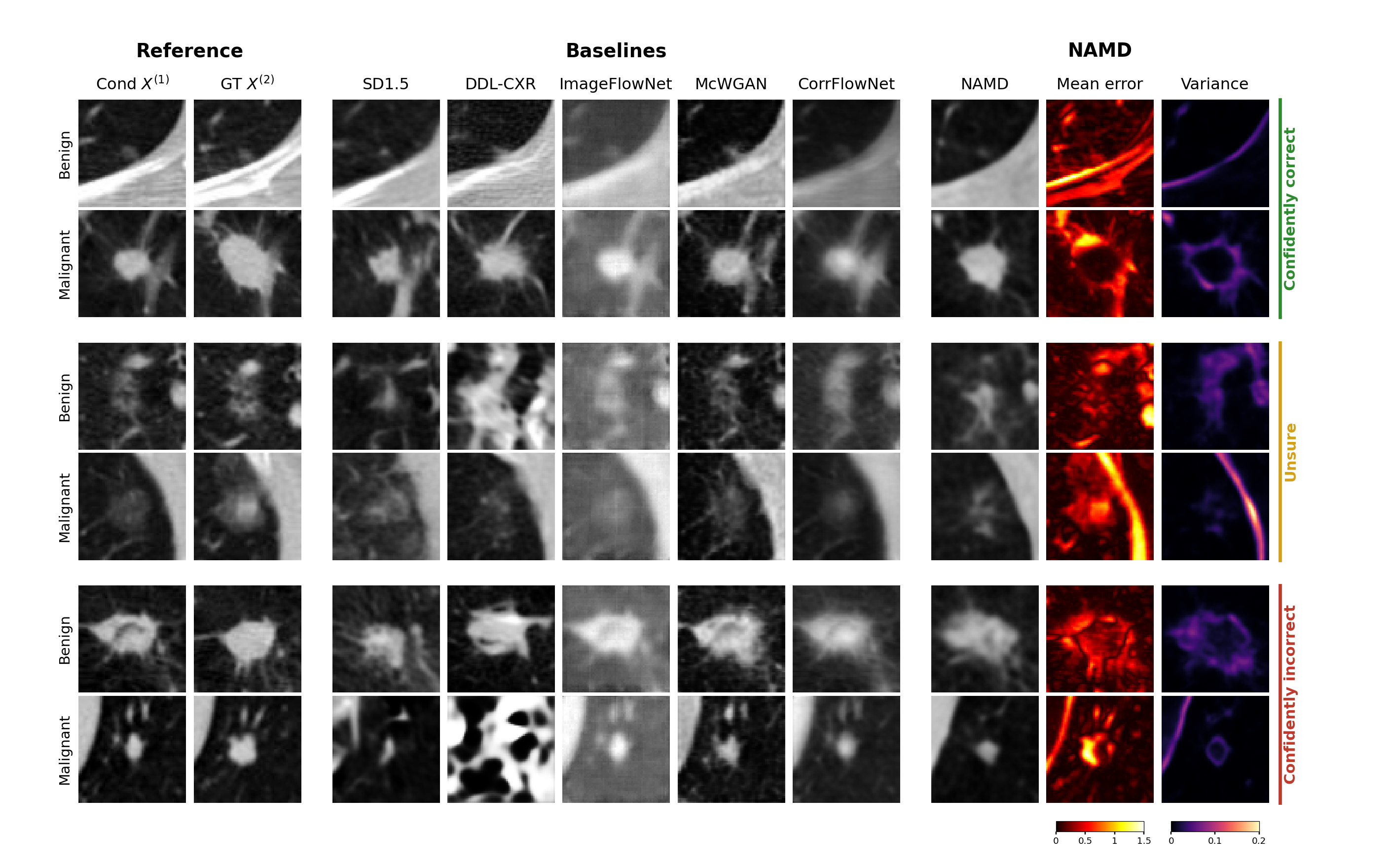}
    \caption{Generated follow-up examples across confidence levels
    (confidently correct, unsure, confidently incorrect) on NAMD, one benign and one malignant each, alongside baseline method samples. Pixel-wise variance and mean-error maps are across 20 runs.}
    \vspace{-18pt}
    \label{fig:qualitative_examples}
\end{figure*}

%% file: Sections/06_Conclusion.tex
\section{Conclusion}
We introduce Nodule-Aligned Multimodal Diffusion (NAMD), a generative framework for predicting longitudinal nodule progression that integrates a nodule-aligned latent space with an LLM-driven conditional diffusion backbone. Our experiments show that NAMD generates high-quality, clinically meaningful follow-up images with diagnostic utility comparable to real follow-ups and outperforms state-of-the-art baselines, while remaining competitive in image quality and enabling early lung cancer diagnosis.

%% file: Sections/Appendix.tex
\section{Training Details}

All NAMD training runs use the AdamW optimizer~\citep{loshchilov2019decoupledweightdecayregularization}. We adopt the VAE and U-Net architectures, along with pretrained weights, from Stable Diffusion v1.5~\citep{Rombach_2022_CVPR}. The VAE has a latent dimension of 4 and a spatial compression factor of 8. The U-Net backbone follows the standard Stable Diffusion configuration, with 320 base channels, channel multipliers of \texttt{[1, 2, 4, 4]}, two residual blocks per resolution, and eight attention heads.

We first fine-tune the VAE with a learning rate of $5 \times 10^{-5}$ and a batch size of 64. The U-Net is then fine-tuned for unconditional generation using a linear learning-rate warmup from $1 \times 10^{-6}$ to $1 \times 10^{-4}$ over 4,000 iterations, followed by cosine decay to a minimum learning rate of $1 \times 10^{-5}$ over 100,000 steps, with a batch size of 16. Subsequently, the U-Net is further fine-tuned for conditional generation with a learning rate of $2 \times 10^{-5}$ and a batch size of 8. We employ hybrid conditioning by concatenating condition image latents with the noisy latent, resulting in eight input channels, and by applying cross-attention with text embeddings extracted from MedGemma 1.5 4B (context dimension 2560). The unconditional diffusion process uses 1,000 timesteps with a linear noise schedule ranging from $\beta_1 = 8.5 \times 10^{-4}$ to $\beta_T = 1.2 \times 10^{-2}$. Training is conducted for 60 epochs, with gradient clipping (maximum norm = 1.0), deterministic operations for reproducibility (seed = 23), and 32-bit floating-point precision.
\section{EHR Information and LLM Template}
This section presents details on the 13 EHR features corresponding to each lung LDCT image as detailed in Table~\ref{tab:nodule_features}, as well as an example of a prompt that contains the feature information fed to the LLM later as detailed in Figure~\ref{fig:generated_report}.

\begin{table*}[h]
\centering
\fontsize{9}{10}\selectfont
\renewcommand{\arraystretch}{1.3} 
\caption{Description of the 13 features containing nodule and patient-level EHR information.}
\begin{tabular}{@{}l l l p{6cm}@{}}
\toprule
& \textbf{Feature Name} & \textbf{Description} & \textbf{Value and Units} \\ 
\midrule
\ldelim\{{5}{*}[\textbf{Nodule}] 
& SCT\_PRE\_ATT & Predominant attenuation & \textcolor{Violet}{Soft, Ground Glass, Part Solid} \\
& SCT\_EPI\_LOC & Location of nodule in the lung & \textcolor{Violet}{Right Upper/Middle/Lower lobe, Left Upper/Lower Lobe, Lingula} \\
& SCT\_LONG\_DIA & Longest diameter & \textcolor{RoyalBlue}{Millimeters} \\
& SCT\_PERP\_DIA & Perpendicular diameter & \textcolor{RoyalBlue}{Millimeters} \\
& SCT\_MARGINS & Margin of the nodule & \textcolor{Violet}{Spiculated, Smooth, Poorly Defined} \\

\midrule

\ldelim\{{8}{*}[\textbf{Patient}] 
& age & Age & \textcolor{RoyalBlue}{Years} \\
& diagemph & Diagnosis to Emphysema & \textcolor{ForestGreen}{Yes/No} \\
& gender & Gender & \textcolor{ForestGreen}{Male/Female} \\
& famfather & Family history, Father & \textcolor{ForestGreen}{Yes/No} \\
& fammother & Family history, Mother & \textcolor{ForestGreen}{Yes/No} \\
& fambrother & Family history, Brother & \textcolor{ForestGreen}{Yes/No} \\
& famsister & Family history, Sister & \textcolor{ForestGreen}{Yes/No} \\
& famchild & Family history, Child & \textcolor{ForestGreen}{Yes/No} \\
\midrule
\multicolumn{4}{l}{\small \textbf{Legend:} \textcolor{RoyalBlue}{Continuous Variable} $\cdot$ \textcolor{Violet}{Multi-Category Variable} $\cdot$ \textcolor{ForestGreen}{Binary Variable}} \\
\bottomrule
\end{tabular}

\label{tab:nodule_features}
\end{table*}
\begin{figure}[ht]
    \centering
    \begin{tcolorbox}[colback=gray!5, colframe=black!75, title=\textbf{Generated Report Example}]
        Lung cancer screening with low dose computed tomography performed for this \textbf{68} years old \textbf{male} \textbf{with prior diagnosis of emphysema} \textbf{and family history of cancer}. A \textbf{part solid} nodule, with \textbf{spiculated margin}, \textbf{27} mm in size is present in the \textbf{right lower lobe}.
    \end{tcolorbox}
    
    \caption{An example of a natural language prompt generated from the input features.}
    \label{fig:generated_report}
\end{figure}

\section{Prediction Variance across runs}
\label{app:pred_var}

\begin{figure}[ht]
    \centering
    \includegraphics[width=\linewidth]{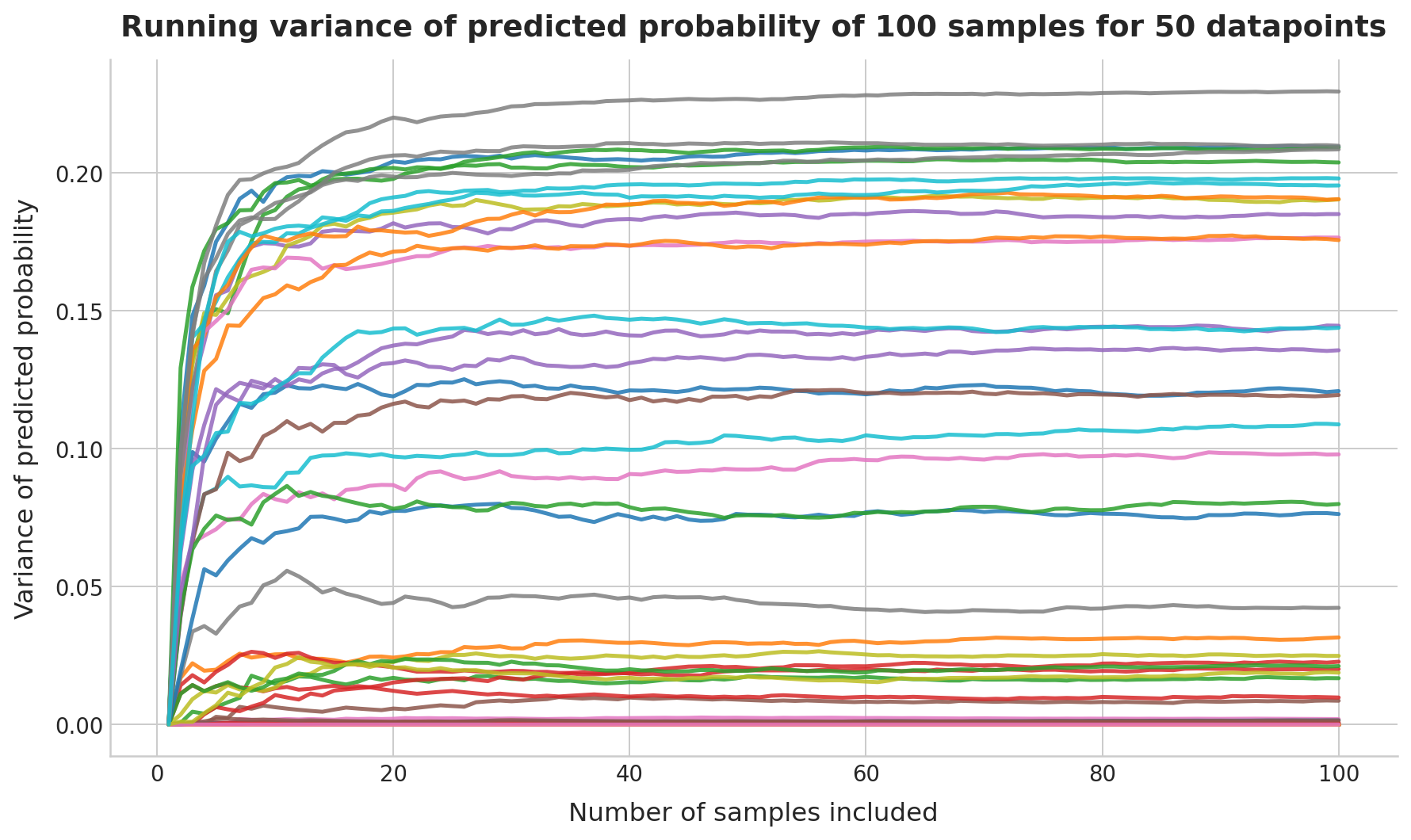}
    \caption{Running variance of ViT prediction probabilities across $K$ samples from the diffusion model for 50 datapoints. Each curve represents one datapoint. }
    \label{fig:rolling_variance}
\end{figure}
Due to the stochastic nature of the diffusion sampling process via DDIM, the generated images will be slightly different with different initial noise. When these generated samples are evaluated by our downstream Vision Transformer (ViT), this diversity naturally translates into fluctuations in the predicted prxobabilities, which in turn causes variations in evaluation metrics in AUROC and AUPRC. To determine an optimal $K$ value that balances evaluation stability with computational efficiency, we graph the running variance for 50 randomly sampled datapoints in Figure~\ref{fig:rolling_variance} and find the minimum $K$ where prediction variance stabilizes. As observed, the running variance for the majority of datapoints initially rises but plateaus after approximately $K=20$ samples. We therefore select $K=20$ for evaluation. Every reported metric is an average across 20 samples. 

\section{Ablation on Pretrained Weights}
\label{app:pretrained_ablation}
\begin{table*}[h]
\centering
\caption{Effect of SD1.5 initialization (VAE + U-Net) on downstream utility and image fidelity.}
\label{tab:pretrain_ablation}
\begin{tabular}{lcccc}
\toprule
Initialization & LPIPS $\downarrow$ & FID $\downarrow$ & AUROC $\uparrow$ & AUPRC $\uparrow$   \\
\midrule
From scratch       & \textbf{0.206 $\pm$ 0.002} & \textbf{77.290 $\pm$ 0.811} & 0.767 $\pm$ 0.026 & 0.305 $\pm$ 0.040 \\
Pretrained (SD1.5) & 0.220 $\pm$ 0.001 & 82.973 $\pm$ 0.691 &  \textbf{0.805 $\pm$ 0.018} &  \textbf{0.347 $\pm$ 0.029} \\
\bottomrule
\end{tabular}
\end{table*}
Table~\ref{tab:pretrain_ablation} isolates the effect of SD1.5 pretrained weight initialization used in NAMD training for the VAE and UNet backbone. Initializing from scratch (with the same architecture) yields better perceptual and distributional fidelity (LPIPS 0.206 vs 0.220, FID 77.29 vs 82.97). The natural-image prior of SD1.5 is distributionally distant from LDCT images, so training from scratch can lead to better image quality. On the other hand, SD1.5 initialization leads to higher performances in diagnosis (AUROC 0.805 vs 0.767, AUPRC 0.347 vs 0.305). As pretrained weights supply a well-conditioned initialization that already has structure compared to randomly initialized weights, nodule-alignment and LLM conditioning in NAMD can be more easily imposed, which can lead to better diagnostic performance. 

%% file: main.bib
@String(CVPR= {IEEE Conf. Comput. Vis. Pattern Recog.})

@String(ICCV= {Int. Conf. Comput. Vis.})

@String(ICASSP=	{ICASSP})

@String(ICLR = {Int. Conf. Learn. Represent.})

@String(CVPR  = {CVPR})

@String(ICCV  = {ICCV})

@String(ICLR  = {ICLR})

@inproceedings{zhang2018perceptual,
  title={The Unreasonable Effectiveness of Deep Features as a Perceptual Metric},
  author={Zhang, Richard and Isola, Phillip and Efros, Alexei A and Shechtman, Eli and Wang, Oliver},
  booktitle={CVPR},
  year={2018}
}

@inproceedings{DBLP:journals/corr/KingmaW13,
  author       = {Diederik P. Kingma and
                  Max Welling},
  editor       = {Yoshua Bengio and
                  Yann LeCun},
  title        = {Auto-Encoding Variational Bayes},
  booktitle    = {2nd International Conference on Learning Representations, {ICLR} 2014,
                  Banff, AB, Canada, April 14-16, 2014, Conference Track Proceedings},
  year         = {2014},
  url          = {http://arxiv.org/abs/1312.6114},
  bibsource    = {dblp computer science bibliography, https://dblp.org}
}

@InProceedings{Rombach_2022_CVPR,
    author    = {Rombach, Robin and Blattmann, Andreas and Lorenz, Dominik and Esser, Patrick and Ommer, Bj\"orn},
    title     = {High-Resolution Image Synthesis With Latent Diffusion Models},
    booktitle = {Proceedings of the IEEE/CVF Conference on Computer Vision and Pattern Recognition (CVPR)},
    month     = {June},
    year      = {2022},
    pages     = {10684-10695}
}

@inproceedings{devlin-etal-2019-bert,
    title = "{BERT}: Pre-training of Deep Bidirectional Transformers for Language Understanding",
    author = "Devlin, Jacob  and
      Chang, Ming-Wei  and
      Lee, Kenton  and
      Toutanova, Kristina",
    editor = "Burstein, Jill  and
      Doran, Christy  and
      Solorio, Thamar",
    booktitle = "Proceedings of the 2019 Conference of the North {A}merican Chapter of the Association for Computational Linguistics: Human Language Technologies, Volume 1 (Long and Short Papers)",
    month = jun,
    year = "2019",
    address = "Minneapolis, Minnesota",
    publisher = "Association for Computational Linguistics",
    url = "https://aclanthology.org/N19-1423/",
    doi = "10.18653/v1/N19-1423",
    pages = "4171--4186"
}

@InProceedings{pmlr-v139-radford21a,
  title = 	 {Learning Transferable Visual Models From Natural Language Supervision},
  author =       {Radford, Alec and Kim, Jong Wook and Hallacy, Chris and Ramesh, Aditya and Goh, Gabriel and Agarwal, Sandhini and Sastry, Girish and Askell, Amanda and Mishkin, Pamela and Clark, Jack and Krueger, Gretchen and Sutskever, Ilya},
  booktitle = 	 {Proceedings of the 38th International Conference on Machine Learning},
  pages = 	 {8748--8763},
  year = 	 {2021},
  editor = 	 {Meila, Marina and Zhang, Tong},
  volume = 	 {139},
  series = 	 {Proceedings of Machine Learning Research},
  month = 	 {18--24 Jul},
  publisher =    {PMLR},
  url = 	 {https://proceedings.mlr.press/v139/radford21a.html}
}

@inproceedings{
yao2024addressing,
title={Addressing Asynchronicity in Clinical Multimodal Fusion via Individualized Chest X-ray Generation},
author={Wenfang Yao and Chen Liu and Kejing Yin and William K. Cheung and Jing Qin},
booktitle={The Thirty-eighth Annual Conference on Neural Information Processing Systems},
year={2024},
url={https://openreview.net/forum?id=uCvdw0IOuU}
}

@article{oh2024llm,
  title={LLM-driven multimodal target volume contouring in radiation oncology},
  author={Oh, Yujin and Park, Sangjoon and Byun, Hwa Kyung and Cho, Yeona and Lee, Ik Jae and Kim, Jin Sung and Ye, Jong Chul},
  journal={Nature Communications},
  volume={15},
  number={1},
  pages={9186},
  year={2024},
  publisher={Nature Publishing Group UK London}
}

@article{national2011national,
  title={The national lung screening trial: overview and study design},
  author={National Lung Screening Trial Research Team},
  journal={Radiology},
  volume={258},
  number={1},
  pages={243--253},
  year={2011},
  publisher={Radiological Society of North America, Inc.}
}

@misc{flux2024,
    author={Black Forest Labs},
    title={FLUX},
    year={2024},
    howpublished={\url{https://github.com/black-forest-labs/flux}},
}

@inproceedings{liu2025imageflownet,
  title={ImageFlowNet: Forecasting Multiscale Image-Level Trajectories of Disease Progression with Irregularly-Sampled Longitudinal Medical Images},
  author={Liu, Chen and Xu, Ke and Shen, Liangbo L and Huguet, Guillaume and Wang, Zilong and Tong, Alexander and Bzdok, Danilo and Stewart, Jay and Wang, Jay C and Del Priore, Lucian V and Krishnaswamy, Smita},
  booktitle={ICASSP 2025-2025 IEEE International Conference on Acoustics, Speech and Signal Processing (ICASSP)},
  year={2025},
  organization={IEEE}
}

@online{Cancer,
  title     = {Cancer},
  author    = {World Health Organization WHO},
  year      = {2025},
  url       = {https://www.who.int/news-room/fact-sheets/detail/cancer#:~:text=schools%20and%20workplaces).-,Early%20detection,and%20the%20majority%20of%20cancers},
  urldate   = {2026-01-31}
}

@online{CancerStat,
  title     = {Cancer Stat Facts: Lung and Bronchus Cancer},
  author    = {National Cancer Institute},
  year      = {2025},
  url       = {https://seer.cancer.gov/statfacts/html/lungb.html},
  urldate   = {2026-01-31}
}

@article{crosby2022early,
  title={Early detection of cancer},
  author={Crosby, David and Bhatia, Sangeeta and Brindle, Kevin M and Coussens, Lisa M and Dive, Caroline and Emberton, Mark and Esener, Sadik and Fitzgerald, Rebecca C and Gambhir, Sanjiv S and Kuhn, Peter and others},
  journal={Science},
  volume={375},
  number={6586},
  pages={eaay9040},
  year={2022},
  publisher={American Association for the Advancement of Science}
}

@online{Assessment,
  title     = {Early-Stage Lung Cancer: Assessment and Treatment},
  author = {Kim, E. S. and Benbow, J.},
  year      = {2025},
  url       = {https://www.healio.com/clinical-guidance/genomics/early-stage-lung-cancer-treatment},
  urldate   = {2026-01-31}
}

@InProceedings{Zhang_2023_ICCV,
    author    = {Zhang, Lvmin and Rao, Anyi and Agrawala, Maneesh},
    title     = {Adding Conditional Control to Text-to-Image Diffusion Models},
    booktitle = {Proceedings of the IEEE/CVF International Conference on Computer Vision (ICCV)},
    month     = {October},
    year      = {2023},
    pages     = {3836-3847}
}

@InProceedings{Tan_2025_ICCV,
    author    = {Tan, Zhenxiong and Liu, Songhua and Yang, Xingyi and Xue, Qiaochu and Wang, Xinchao},
    title     = {OminiControl: Minimal and Universal Control for Diffusion Transformer},
    booktitle = {Proceedings of the IEEE/CVF International Conference on Computer Vision (ICCV)},
    month     = {October},
    year      = {2025},
    pages     = {14940-14950}
}

@misc{ho2020denoisingdiffusionprobabilisticmodels,
      title={Denoising Diffusion Probabilistic Models}, 
      author={Jonathan Ho and Ajay Jain and Pieter Abbeel},
      year={2020},
      eprint={2006.11239},
      archivePrefix={arXiv},
      primaryClass={cs.LG},
      url={https://arxiv.org/abs/2006.11239}, 
}

@misc{sohldickstein2015deepunsupervisedlearningusing,
      title={Deep Unsupervised Learning using Nonequilibrium Thermodynamics}, 
      author={Jascha Sohl-Dickstein and Eric A. Weiss and Niru Maheswaranathan and Surya Ganguli},
      year={2015},
      eprint={1503.03585},
      archivePrefix={arXiv},
      primaryClass={cs.LG},
      url={https://arxiv.org/abs/1503.03585}, 
}

@article{mou2023t2i,
  title={T2i-adapter: Learning adapters to dig out more controllable ability for text-to-image diffusion models},
  author={Mou, Chong and Wang, Xintao and Xie, Liangbin and Wu, Yanze and Zhang, Jian and Qi, Zhongang and Shan, Ying and Qie, Xiaohu},
  journal={arXiv preprint arXiv:2302.08453},
  year={2023}
}

@InProceedings{Peebles_2023_ICCV,
    author    = {Peebles, William and Xie, Saining},
    title     = {Scalable Diffusion Models with Transformers},
    booktitle = {Proceedings of the IEEE/CVF International Conference on Computer Vision (ICCV)},
    month     = {October},
    year      = {2023},
    pages     = {4195-4205}
}

@inproceedings{weber2023cascaded,
  title={Cascaded Latent Diffusion Models for High-Resolution Chest X-ray Synthesis},
  author={Weber, Tobias and Ingrisch, Michael and Bischl, Bernd and R{\"u}gamer, David},
  booktitle={Advances in Knowledge Discovery and Data Mining: 27th Pacific-Asia Conference, PAKDD 2023},
  year={2023},
  organization={Springer}
}

@InProceedings{AroMeh_CXRTFT_MICCAI2025,
        author = { Arora, Mehak AND Ali, Ayman AND Wu, Kaiyuan AND Davis, Carolyn AND Shimazui, Takashi AND Alwakeel, Mahmoud AND Moas, Victor AND Yang, Philip AND Esper, Annette AND Kamaleswaran, Rishikesan},
        title = { { CXR-TFT: Multi-Modal Temporal Fusion Transformer for Predicting Chest X-ray Trajectories } },
        booktitle = {proceedings of Medical Image Computing and Computer Assisted Intervention -- MICCAI 2025},
        year = {2025},
        publisher = {Springer Nature Switzerland},
        volume = {LNCS 15974},
        month = {September},
        page = {158 -- 166}
}

@inproceedings{
chen2026learning,
title={Learning Patient-Specific Disease Dynamics With Latent Flow Matching For Longitudinal Imaging Generation},
author={Hao Chen and Rui Yin and Yifan Chen and Qi Chen and Chao Li},
booktitle={The Fourteenth International Conference on Learning Representations},
year={2026},
url={https://openreview.net/forum?id=cuGnuOfQ4U}
}

@article{wang2024enhancing,
  title={Enhancing early lung cancer diagnosis: predicting Lung Nodule progression in follow-up low-dose CT scan with deep generative model},
  author={Wang, Yifan and Zhou, Chuan and Ying, Lei and Chan, Heang-Ping and Lee, Elizabeth and Chughtai, Aamer and Hadjiiski, Lubomir M and Kazerooni, Ella A},
  journal={Cancers},
  volume={16},
  number={12},
  pages={2229},
  year={2024},
  publisher={MDPI}
}

@article{wu2025early,
  title={Early Lung Cancer Diagnosis from Virtual Follow-up LDCT Generation via Correlational Autoencoder and Latent Flow Matching},
  author={Wu, Yutong and Wang, Yifan and Zhang, Qining and Zhou, Chuan and Ying, Lei},
  journal={arXiv preprint arXiv:2511.18185},
  year={2025}
}

@article{sellergren2025medgemma,
  title={MedGemma Technical Report},
  author={Sellergren, Andrew and Kazemzadeh, Sahar and Jaroensri, Tiam and Kiraly, Atilla and Traverse, Madeleine and Kohlberger, Timo and Xu, Shawn and Jamil, Fayaz and Hughes, Cían and Lau, Charles and others},
  journal={arXiv preprint arXiv:2507.05201},
  year={2025}
}

@article{JMLR:v9:vandermaaten08a,
  author  = {Laurens van der Maaten and Geoffrey Hinton},
  title   = {Visualizing Data using t-SNE},
  journal = {Journal of Machine Learning Research},
  year    = {2008},
  volume  = {9},
  number  = {86},
  pages   = {2579--2605},
  url     = {http://jmlr.org/papers/v9/vandermaaten08a.html}
}

@misc{loshchilov2019decoupledweightdecayregularization,
      title={Decoupled Weight Decay Regularization}, 
      author={Ilya Loshchilov and Frank Hutter},
      year={2019},
      eprint={1711.05101},
      archivePrefix={arXiv},
      primaryClass={cs.LG},
      url={https://arxiv.org/abs/1711.05101}, 
}

@article{gupta2024texture,
  title={Texture and radiomics inspired data-driven cancerous lung nodules severity classification},
  author={Gupta, Himanshu and Singh, Himanshu and Kumar, Anil},
  journal={Biomedical Signal Processing and Control},
  volume={88},
  pages={105543},
  year={2024},
  publisher={Elsevier}
}

@article{liu2024lung,
  title={Lung nodule classification using radiomics model trained on degraded SDCT images},
  author={Liu, Jiaying and Corti, Anna and Corino, Valentina DA and Mainardi, Luca},
  journal={Computer Methods and Programs in Biomedicine},
  volume={257},
  pages={108474},
  year={2024},
  publisher={Elsevier}
}

@article{yu2025etmo,
  title={ETMO-NAS: An efficient two-step multimodal one-shot NAS for lung nodules classification},
  author={Yu, Jiandong and Li, Tongtong and Shi, Xuerong and Zhao, Ziyang and Chen, Miao and Zhang, Yu and Wang, Junyu and Yao, Zhijun and Fang, Lei and Hu, Bin},
  journal={Biomedical Signal Processing and Control},
  volume={104},
  pages={107479},
  year={2025},
  publisher={Elsevier}
}

@article{ardila2019end,
  title={End-to-end lung cancer screening with three-dimensional deep learning on low-dose chest computed tomography},
  author={Ardila, Diego and Kiraly, Atilla P and others},
  journal={Nature medicine},
  volume={25},
  number={6},
  pages={954--961},
  year={2019},
  publisher={Nature Publishing Group US New York}
}

@article{wang2024leveraging,
  title={Leveraging serial low-dose CT scans in radiomics-based reinforcement learning to improve early diagnosis of lung cancer at baseline screening},
  author={Wang, Yifan and Zhou, Chuan and Ying, Lei and Lee, Elizabeth and Chan, Heang-Ping and Chughtai, Aamer and Hadjiiski, Lubomir M and Kazerooni, Ella A},
  journal={Radiology: Cardiothoracic Imaging},
  volume={6},
  number={3},
  pages={e230196},
  year={2024},
  publisher={Radiological Society of North America}
}

@INPROCEEDINGS {TangICCV2025TULIP,
author = { Tang, Zineng and Lian, Long and Eisape, Seun and Wang, XuDong and Herzig, Roei and Yala, Adam and Suhr, Alane and Darrell, Trevor and Chan, David M. },
booktitle = { 2025 IEEE/CVF International Conference on Computer Vision Workshops (ICCVW) },
title = {{ TULIP: Contrastive Image-Text Learning with Richer Vision Understanding }},
year = {2025},
volume = {},
ISSN = {},
pages = {4326-4336},
doi = {10.1109/ICCVW69036.2025.00449},
url = {https://doi.ieeecomputersociety.org/10.1109/ICCVW69036.2025.00449},
publisher = {IEEE Computer Society},
address = {Los Alamitos, CA, USA},
month =Oct}

@inproceedings{liu-etal-2025-medebench,
    title = "{M}ed{EB}ench: Diagnosing Reliability in Text-Guided Medical Image Editing",
    author = "Liu, Minghao  and
      He, Zhitao  and
      Fan, Zhiyuan  and
      Wang, Qingyun  and
      Fung, Yi R.",
    editor = "Christodoulopoulos, Christos  and
      Chakraborty, Tanmoy  and
      Rose, Carolyn  and
      Peng, Violet",
    booktitle = "Findings of the Association for Computational Linguistics: EMNLP 2025",
    month = nov,
    year = "2025",
    address = "Suzhou, China",
    publisher = "Association for Computational Linguistics",
    url = "https://aclanthology.org/2025.findings-emnlp.41/",
    doi = "10.18653/v1/2025.findings-emnlp.41",
    pages = "767--791",
    ISBN = "979-8-89176-335-7"
}

@inproceedings{
zheng2026diffusion,
title={Diffusion Transformers with Representation Autoencoders},
author={Boyang Zheng and Nanye Ma and Shengbang Tong and Saining Xie},
booktitle={The Fourteenth International Conference on Learning Representations},
year={2026},
url={https://openreview.net/forum?id=0u1LigJaab}
}

@inproceedings{
song2021denoising,
title={Denoising Diffusion Implicit Models},
author={Jiaming Song and Chenlin Meng and Stefano Ermon},
booktitle={International Conference on Learning Representations},
year={2021},
url={https://openreview.net/forum?id=St1giarCHLP}
}

@inproceedings{supcon,
 author = {Khosla, Prannay and Teterwak, Piotr and Wang, Chen and Sarna, Aaron and Tian, Yonglong and Isola, Phillip and Maschinot, Aaron and Liu, Ce and Krishnan, Dilip},
 booktitle = {Advances in Neural Information Processing Systems},
 editor = {H. Larochelle and M. Ranzato and R. Hadsell and M.F. Balcan and H. Lin},
 pages = {18661--18673},
 publisher = {Curran Associates, Inc.},
 title = {Supervised Contrastive Learning},
 url = {https://proceedings.neurips.cc/paper_files/paper/2020/file/d89a66c7c80a29b1bdbab0f2a1a94af8-Paper.pdf},
 volume = {33},
 year = {2020}
}

@inproceedings{wang-etal-2022-medclip,
    title = "{M}ed{CLIP}: Contrastive Learning from Unpaired Medical Images and Text",
    author = "Wang, Zifeng  and
      Wu, Zhenbang  and
      Agarwal, Dinesh  and
      Sun, Jimeng",
    editor = "Goldberg, Yoav  and
      Kozareva, Zornitsa  and
      Zhang, Yue",
    booktitle = "Proceedings of the 2022 Conference on Empirical Methods in Natural Language Processing",
    month = dec,
    year = "2022",
    address = "Abu Dhabi, United Arab Emirates",
    publisher = "Association for Computational Linguistics",
    url = "https://aclanthology.org/2022.emnlp-main.256/",
    doi = "10.18653/v1/2022.emnlp-main.256",
    pages = "3876--3887"
}

@article{wang2025duke,
  title={The Duke Lung Cancer Screening (DLCS) dataset: a reference dataset of annotated low-dose screening thoracic CT},
  author={Wang, Avivah J and Tushar, Fakrul Islam and Harowicz, Michael R and Tong, Betty C and Lafata, Kyle J and Tailor, Tina D and Lo, Joseph Y},
  journal={Radiology: Artificial Intelligence},
  volume={7},
  number={4},
  pages={e240248},
  year={2025},
  publisher={Radiological Society of North America}
}

@misc{peeters_2025_15094631,
  author       = {Peeters, Dré and
                  Obreja, Bogdan and
                  Antonissen, Noa and
                  Jacobs, Colin},
  title        = {Benchmarking of Artificial Intelligence and
                   Radiologists for Lung Cancer Screening in CT: The
                   LUNA25 Challenge
                  },
  month        = mar,
  year         = 2025,
  publisher    = {Zenodo},
  doi          = {10.5281/zenodo.15094631},
  url          = {https://doi.org/10.5281/zenodo.15094631},
}

@article{setio2017validation,
  title={Validation, comparison, and combination of algorithms for automatic detection of pulmonary nodules in computed tomography images: the LUNA16 challenge},
  author={Setio, Arnaud Arindra Adiyoso and Traverso, Alberto and De Bel, Thomas and Berens, Moira SN and Van Den Bogaard, Cas and Cerello, Piergiorgio and Chen, Hao and Dou, Qi and Fantacci, Maria Evelina and Geurts, Bram and others},
  journal={Medical image analysis},
  volume={42},
  pages={1--13},
  year={2017},
  publisher={Elsevier}
}

@article{ma2024segment,
  title={Segment anything in medical images},
  author={Ma, Jun and He, Yuting and Li, Feifei and Han, Lin and You, Chenyu and Wang, Bo},
  journal={Nature communications},
  volume={15},
  number={1},
  pages={654},
  year={2024},
  publisher={Nature Publishing Group UK London}
}

@article{hu2024ella,
  title={Ella: Equip diffusion models with llm for enhanced semantic alignment},
  author={Hu, Xiwei and Wang, Rui and Fang, Yixiao and Fu, Bin and Cheng, Pei and Yu, Gang},
  journal={arXiv preprint arXiv:2403.05135},
  year={2024}
}

@inproceedings{yang2024mastering,
  title={Mastering Text-to-Image Diffusion: Recaptioning, Planning, and Generating with Multimodal LLMs},
  author={Yang, Ling and Yu, Zhaochen and Meng, Chenlin and Xu, Minkai and Ermon, Stefano and Cui, Bin},
  booktitle={International Conference on Machine Learning},
  year={2024}
}
